\documentclass[11pt, a4paper]{lumia}

\usepackage[sort&compress]{natbib}
\newcommand{\gray}[1]{\textcolor{gray}{#1}}

\usepackage{xspace}

\theoremstyle{plain}

\newtheorem*{proposition*}{Proposition}

\theoremstyle{definition}

\theoremstyle{definition}

\def\eqref#1{equation~\ref{#1}}

\usepackage{graphicx}           
\usepackage{tikz}               
\usepackage[edges]{forest}      

\usepackage{url}                
\usepackage{xurl}               

\usepackage{array}              
\usepackage{longtable}          
\usepackage{multirow}           
\usepackage{makecell}           
\usepackage{ragged2e}           

\usepackage{mathtools}          
\usepackage{nicefrac}           

\usepackage{algorithm}          
\usepackage{algorithmicx}       
\usepackage{algpseudocode}      
\usepackage{listings}           

\usepackage{subcaption}         
\usepackage{wrapfig}            
\usepackage[export]{adjustbox}  

\usepackage[commandnameprefix=ifneeded]{changes}            
\usepackage{xspace}             
\usepackage[normalem]{ulem}     
\usepackage{CJKutf8}            

\usepackage[tikz]{bclogo}       
\usepackage[framemethod=tikz]{mdframed} 

\usepackage{lipsum}             
\usepackage{tocloft}            
\usepackage{afterpage}          
\usepackage{bbding}             
\usepackage{epigraph}           
\usepackage{minitoc}            
\usepackage{multicol}           
\usepackage{textgreek}          

\newcolumntype{P}[1]{>{\RaggedRight\arraybackslash}p{#1}}

\definecolor{uclablue}{RGB}{39, 116, 174}
\definecolor{bigaired}{RGB}{156, 0, 0}
\definecolor{myblue}{HTML}{598BE7}
\definecolor{mildblue}{RGB}{31,119,180}
\definecolor{sectionblue}{RGB}{70, 130, 180}
\definecolor{methodblue}{RGB}{0, 150, 136}
\definecolor{bgblue}{RGB}{245,243,253}
\definecolor{ttblue}{RGB}{91,194,224}
\definecolor{mygreen}{rgb}{0.64, 0.56, 0.88}
\definecolor{myyellow}{rgb}{0.68, 0.6, 0.1}
\definecolor{fancygreen}{rgb}{0.33, 0.68, 0.20}
\definecolor{salmon}{rgb}{0.94, 0.52, 0.49}
\definecolor{tablegreen}{rgb}{0.82, 0.94, 0.75}
\definecolor{tableblue}{rgb}{0.81, 0.90, 0.94}
\definecolor{tablered}{rgb}{0.97, 0.85, 0.85}
\definecolor{tableorange}{rgb}{0.96, 0.85, 0.81}
\definecolor{myorange}{rgb}{1.0, 0.49, 0.0}
\definecolor{tlgreen}{rgb}{0.33, 0.68, 0.20}
\definecolor{darkgreen}{RGB}{0,100,0}
\definecolor{darkred}{RGB}{200, 0, 0}
\definecolor{customyellow}{HTML}{FFFACD}
\definecolor{refinegreen}{RGB}{0, 128, 75}
\definecolor{scoregreen}{RGB}{34, 139, 34}
\definecolor{hidden-blue}{RGB}{194,232,247}
\definecolor{hidden-black}{RGB}{20,68,106}
\definecolor{yes}{HTML}{C6EFCE}
\definecolor{no}{HTML}{FFC7CE}
\definecolor{partial}{HTML}{FFEB9C}
\definecolor{external}{HTML}{D9E1F2}
\definecolor{hdr}{HTML}{F2F2F2}
\definecolor{GRPOrow}{gray}{0.96}
\definecolor{FlowRLrow}{RGB}{225,236,255}
\definecolor{FlowBlue}{RGB}{80,120,210}
\definecolor{GRPOGray}{gray}{0.35}

\hypersetup{
    colorlinks=true, 
    citecolor=uclablue, 
    linkcolor=bigaired,
    urlcolor=darkblue
}

\setlist[itemize]{leftmargin=20pt, noitemsep, topsep=0pt}

\NewDocumentCommand{\kaiyan}{mO{}}{\textcolor{purple}{\textsuperscript{\textit{kaiyan}}\textsf{\textbf{\small[#1]}}}}
\NewDocumentCommand{\yuxin}{mO{}}{\textcolor{cyan}{\textsuperscript{\textit{yuxin}}\textsf{\textbf{\small[#1]}}}}
\NewDocumentCommand{\bx}{mO{}}{\textcolor{green}{\textsuperscript{\textit{bx}}\textsf{\textbf{\small[#1]}}}}
\NewDocumentCommand{\at}{mO{}}{\textcolor{red}{\textsuperscript{\textit{AT}}\textsf{\textbf{\small[#1]}}}}
\NewDocumentCommand{\re}{mO{}}{\textcolor{blue}{\textsuperscript{\textit{RE}}\textsf{\textbf{\small[#1]}}}}
\NewDocumentCommand{\ybsun}{mO{}}{\textcolor{magenta}{\textsuperscript{\textit{youbang}}\textsf{\textbf{\small[#1]}}}}
\NewDocumentCommand{\runze}{mO{}}{\textcolor{orange}{\textsuperscript{\textit{runze}}\textsf{\textbf{\small[#1]}}}}
\NewDocumentCommand{\add}{mO{}}{\textcolor{darkgreen}{\textsuperscript{\textit{Maybe Consider Discuss}}\textsf{\textbf{[#1]}}}}

\newcommand{\cmark}{\textcolor{darkgreen}{\boldmath$\checkmark$}}
\newcommand{\xmark}{\textcolor{darkred}{\boldmath$\times$}}

\newenvironment{itemize*}%
 {\leftmargini=10pt\begin{itemize}%
  \setlength{\itemsep}{0pt}%
  \setlength{\parskip}{0pt}%
  }%
 {\end{itemize}}

\newenvironment{enumerate*}%
 {\begin{enumerate}%
  \setlength{\itemsep}{0pt}%
  \setlength{\parskip}{0pt}}%
 {\end{enumerate}}

\newcommand{\cellstatus}[1]{%
  \begingroup
  \StrTrim{#1}[\statusval]%
  \IfStrEq{\statusval}{Yes}{\cellcolor{yes}\cmark}{}%
  \IfStrEq{\statusval}{No}{\cellcolor{no}\xmark}{}%
  \IfBeginWith{\statusval}{Yes (}{\cellcolor{yes}\cmark~\textit{\statusval\unskip}}{}%
  \IfStrEq{\statusval}{Partial}{\cellcolor{partial}\textbf{Partial}}{}%
  \IfStrEq{\statusval}{External}{\cellcolor{external}\textbf{External}}{}%
  \endgroup
}

\newtcolorbox{myboxi}[1][]{
  breakable,
  title=#1,
  colback=red!5,
  colbacktitle=red!5,
  coltitle=black,
  fonttitle=\bfseries,
  bottomrule=0pt,
  toprule=0pt,
  leftrule=2pt,
  rightrule=2pt,
  titlerule=0pt,
  arc=0pt,
  outer arc=0pt,
  colframe=red,
}

\newtcolorbox{myboxnote}[1][]{
  breakable,
  title=#1,
  colback=orange!0,
  colbacktitle=orange!0,
  coltitle=black,
  fonttitle=\bfseries,
  bottomrule=0pt,
  toprule=0pt,
  leftrule=2pt,
  rightrule=2pt,
  titlerule=0pt,
  arc=0pt,
  outer arc=0pt,
  colframe=orange,
}

\newtcolorbox{myboxii}[1][]{
  breakable,
  freelance,
  title=#1,
  colback=white,
  colbacktitle=white,
  coltitle=black,
  fonttitle=\bfseries,
  bottomrule=0pt,
  boxrule=0pt,
  colframe=white,
  overlay unbroken and first={
  \draw[red!75!black,line width=3pt]
    ([xshift=5pt]frame.north west) -- 
    (frame.north west) -- 
    (frame.south west);
  \draw[red!75!black,line width=3pt]
    ([xshift=-5pt]frame.north east) -- 
    (frame.north east) -- 
    (frame.south east);
  },
  overlay unbroken app={
  \draw[red!75!black,line width=3pt,line cap=rect]
    (frame.south west) -- 
    ([xshift=5pt]frame.south west);
  \draw[red!75!black,line width=3pt,line cap=rect]
    (frame.south east) -- 
    ([xshift=-5pt]frame.south east);
  },
  overlay middle and last={
  \draw[red!75!black,line width=3pt]
    (frame.north west) -- 
    (frame.south west);
  \draw[red!75!black,line width=3pt]
    (frame.north east) -- 
    (frame.south east);
  },
  overlay last app={
  \draw[red!75!black,line width=3pt,line cap=rect]
    (frame.south west) --
    ([xshift=5pt]frame.south west);
  \draw[red!75!black,line width=3pt,line cap=rect]
    (frame.south east) --
    ([xshift=-5pt]frame.south east);
  },
}

\tcbset{
  takeawaysbox/.style={
    title=Takeaways,
    colback=lightblue!80,
    colframe=black,
    fonttitle=\bfseries\small,
    coltitle=white,
    colbacktitle=black,
    enhanced,
    attach boxed title to top left={xshift=2.5mm,yshift=-2.5mm},
    boxed title style={rounded corners, size=small, colframe=black, colback=black},
    width=\linewidth,
    arc=3.5mm
  }
}

\mdfdefinestyle{mystyle}{%
  rightline=true,
  innerleftmargin=10,
  innerrightmargin=10,
  outerlinewidth=3pt,
  topline=false,
  rightline=true,
  bottomline=false,
  skipabove=\topsep,
  skipbelow=\topsep
}

\tikzset{%
    every node/.style={font=\tiny},
    parent/.style =          {align=center,text width=2cm,rounded corners=3pt, line width=0.3mm, fill=gray!10,draw=gray!80},
    child/.style =           {align=center,text width=2.0cm,rounded corners=3pt, fill=blue!10,draw=blue!80,line width=0.3mm},
    grandchild/.style =      {align=center,text width=2cm,rounded corners=3pt},
    greatgrandchild/.style = {align=center,text width=1.5cm,rounded corners=3pt},
    greatgrandchild2/.style = {align=center,text width=1.5cm,rounded corners=3pt},    
    referenceblock/.style =  {align=center,text width=1.5cm,rounded corners=2pt},
    pretrain/.style =           {align=center,text width=2.0cm,rounded corners=3pt, fill=blue!10,draw=blue!80,line width=0.3mm},   
    pretrain_work/.style =           {align=center, text width=8.5cm,rounded corners=3pt, fill=blue!10,draw=blue!0,line width=0.3mm},  
    template/.style =           {align=center,text width=2.0cm,rounded corners=3pt, fill=red!10,draw=red!80,line width=0.3mm},   
    template_work/.style =           {align=center,text width=8.5cm,rounded corners=3pt, fill=red!10,draw=red!0,line width=0.3mm},    
    answer/.style =           {align=center,text width=2.0cm,rounded corners=3pt, fill= cyan!10,draw= cyan!80,line width=0.3mm},   
    answer_work/.style =           {align=center,text width=8.5cm,rounded corners=3pt, fill= cyan!10,draw= cyan!0,line width=0.3mm},      
    multiple/.style =           {align=center,text width=2.0cm,rounded corners=3pt, fill= orange!10,draw= orange!80,line width=0.3mm},   
    multiple_work/.style =           {align=center,text width=8.5cm,rounded corners=3pt, fill= orange!10,draw= orange!0,line width=0.3mm},        
    tuning/.style =           {align=center,text width=2.0cm,rounded corners=3pt, fill= magenta!10,draw= magenta!80,line width=0.3mm},   
    tuning_work/.style =           {align=center,text width=8.5cm,rounded corners=3pt, fill= magenta!10,draw= magenta!0,line width=0.3mm},          
}

\newcommand{\lstbg}[3][0pt]{{\fboxsep#1\colorbox{#2}{\strut #3}}}

\lstdefinelanguage{diff}{
  basicstyle=\ttfamily\small,
  morecomment=[f][\lstbg{red!20}]-,
  morecomment=[f][\lstbg{green!20}]+,
}

\lstdefinelanguage{diffpython}{
  language=diff,
  morekeywords={def, if, else, for, while, return, import, from, as, class, with, try, except, finally, raise, lambda, and, or, not, in, is, None, True, False},
  morecomment=[l]{\#},
  morestring=[b]",
  morestring=[b]',
}

\setheadertext{LUMIA Lab}

\correspondingemail{\emailicon \{lucky.wangyixuan, lin.zhouhan\}@gmail.com,  \quad $^\ddagger$ Corresponding Author. }

\githublink{https://github.com/LUMIA-Group/VQKV}

\huggingfacelink{https://huggingface.co/LuckyOrz/vqkv\_llama3.1-8B}

\renewcommand{\today}{2026-3-18}

\title{VQKV: High-Fidelity and High-Ratio Cache Compression via Vector-Quantization}
\setheadertitle{VQKV: High-Fidelity and High-Ratio Cache Compression via Vector-Quantization}

\author{
  \Authfont Yixuan Wang$^{1,2,3}$, 
  Qingyu Shi$^{4}$,
  Jiayu Zhou$^{1}$,
  Dianbo Liu$^{5}$,
  Ziwei He$^{2}$,
  Zhouhan Lin$^{1,2,3\ddagger}$\\
  $^1$ LUMIA Lab, School of Artificial Intelligence, Shanghai Jiao Tong University \\
  $^2$ Shanghai Innovation Institute \\
  $^3$ Shanghai Artificial Intelligence Laboratory \\
  $^4$ University of Cambridge \\
  $^5$ National University of Singapore
}

\begin{document}

\begin{abstract}
The growing context length of Large Language Models (LLMs) enlarges the Key-Value (KV) cache, limiting deployment in resource-limited environments. Prior training-free approaches for KV cache compression typically rely on low-rank approximation or scalar quantization, which fail to simultaneously achieve high compression ratios and high reconstruction fidelity. We propose VQKV, a novel, training-free method introducing vector quantization (VQ) to obtain highly compressed KV representations while preserving high model fidelity, allowing for the representation of thousands of floating-point values with just a few integer indices. As a result, VQKV achieves an 82.8\% compression ratio on LLaMA3.1-8B while retaining 98.6\% of the baseline performance on LongBench and enabling 4.3× longer generation length on the same memory footprint.
\end{abstract}

\maketitle


\section{Introduction}
Large Language Models (LLMs) have already found widespread applications in many fields due to their outstanding capabilities. However, the scaling of context length results in continuous growth of the Key-Value (KV) cache, which in turn limits the feasibility of employing large language models in resource-constrained environments. Although approaches such as token eviction \citep{snapkv, h2o, streamingllm}, feature dimension reduction \citep{palu, asvd, mla}, and scalar quantization \citep{kivi, asvd, KVQuant} can effectively alleviate memory usage, they frequently incur a degradation in model performance.

A training-free strategy is particularly appealing because it avoids costly retraining or fine-tuning, and can be immediately deployed across diverse models and checkpoints without modifying their parameters. Among existing approaches, it is difficult to simultaneously remain training-free and achieve high compression ratios. On the one hand, training-free methods like SnapKV \citep{snapkv} and H2O \citep{h2o} can attain extremely high compression ratios by evicting tokens. However, discarding tokens inevitably introduces irreversible information loss. Other training-free methods that compress along the feature dimension of KV cache incur minimal information loss but struggle to achieve high compression ratios: approaches such as Palu \citep{palu}, ASVD \citep{asvd}, and scalar quantization methods \citep{kivi, KVQuant} often have to sacrifice model performance to improve compression. On the other hand, although methods like MLA \citep{mla} and CSKV \citep{cskv} can also achieve high compression ratios, they require additional training of the large language model itself, which not only alters the model’s original capabilities but also limits their general applicability. Therefore, identifying a training-free compression method that achieves both high compression ratios and high reconstruction fidelity is crucial for improving KV cache compression.

Vector quantization (VQ) \citep{vq-vanilla, vqvae} achieves extremely high compression ratios while enabling high-fidelity reconstruction. By jointly quantizing entire vectors into compact integer indices, VQ has been successfully applied in domains such as text retrieval \citep{pqnn, faiss, johnson2019billion} and computer vision \citep{mae, vqvae, vqgan}. However, directly applying VQ to KV cache compression in large language models is challenging. Large language models are highly sensitive to the value of KV cache and even small perturbations in the KV cache can lead to significant performance degradation. Moreover, RoPE \citep{rope} induces heterogeneous frequency characteristics across dimensions, which results in a markedly different representation structure of the key cache, making it difficult to accurately reconstruct the key cache after RoPE transformation.

In this work, we propose VQKV, a novel training-free KV cache compression method based on vector quantization (VQ). VQKV maps high-dimensional cache vectors to a compact set of codebooks and replaces thousands of floating-point values of the full-precision cache with only a few corresponding integer indices. By adopting SimVQ for joint vector-level quantization, VQKV captures intrinsic data structure and progressively refines representations, enabling high-fidelity reconstruction with negligible impact on model performance. Furthermore, the residual design effectively distributes the variations introduced by RoPE across multiple codebooks, allowing VQKV to robustly handle RoPE-rotated key cache values. As a result, VQKV achieves substantial memory savings with a high compression ratio, while maintaining high fidelity and strong model performance across a wide range of downstream tasks.

Extensive evaluations on LLaMA3.1-8B \citep{llama3.1-8b} across the LongBench \citep{longbench} benchmarks demonstrate that our VQKV achieves an 82.8\% discarding ratio while retaining 98.6\% of the baseline performance. Remarkably, on some tasks, VQKV even surpasses the performance of the uncompressed full-cache baseline. Under identical hardware settings, this memory efficiency translates into up to 4.3× longer generation lengths than the original model.

\section{Related Work}
\paragraph{KV Cache Compression} A large body of prior work on KV cache compression can be broadly categorized into three directions: token eviction \citep{streamingllm, h2o, snapkv, pyramidkv}, feature dimension reduction \citep{mla, palu, asvd}, and scalar quantization \citep{kivi, KVQuant, palu}, each facing an inherent trade-off between achieving high compression ratios and preserving model performance.

Token eviction methods \citep{streamingllm, h2o, snapkv, pyramidkv, kai2025freqkv} reduce memory consumption by selectively discarding or structurally organizing KV representations of less critical tokens. By shortening the effective sequence length, these approaches can achieve high compression ratios without modifying the feature dimension. However, token eviction inevitably introduces irreversible information loss, which limits its robustness, particularly for tasks that require long-range context retrieval or precise recall over extended sequences.

Feature dimension reduction approaches \citep{mla, palu, asvd} exploit redundancy within high-dimensional KV vectors by projecting them into a lower-dimensional space, often via low-rank decomposition or learned projections. While this strategy can yield substantial memory savings, achieving strong performance typically requires additional training or continued pretraining of the LLM, as in MLA \citep{mla} and CSKV \citep{cskv}. Even in training-free settings, these methods often struggle to balance compression ratio and fidelity, as aggressive dimensionality reduction can hinder accurate reconstruction of the original KV representations.

Scalar quantization methods \citep{kivi, KVQuant, palu} compress the KV cache by independently reducing the bit-width of each floating-point value, offering a simple and training-free solution with clear memory and computational benefits. Nevertheless, by treating each feature independently, scalar quantization fails to exploit the correlations and structural information within high-dimensional vectors. As a result, maintaining model performance at high compression ratios—especially under very low bit-width settings—remains challenging due to substantial quantization error.

Overall, existing KV cache compression methods exhibit a fundamental tension between achieving high compression ratios and preserving model capability. Token eviction suffers from irreversible information loss, feature dimension reduction often requires additional training or sacrifices fidelity at high compression levels, and scalar quantization struggles to maintain precision when pushed to aggressive compression. This motivates the need for a compression approach that can simultaneously achieve high compression ratios while faithfully preserving the information in the KV cache.

\paragraph{Vector Quantization} Vector quantization is a widely recognized technique known for its effective compression and representation abilities. Wav2Vec \citep{wav2vec2} and HuBERT \citep{hubert} use VQ to extract discrete pseudo labels from continuous raw waves for unsupervised learning. SoundStream \citep{soundstream} introduces residual skill to VQ, thus enriching the representation space of codebooks. In computer vision, VQ is generally used for concentrating information from pixels \citep{mae, vqvae, vqgan, movq}. At the same time, VQ helps introduce information from other modalities into LLMs \citep{lumina-mgpt, speechgpt}. Other works use VQ for constraining the specialized formats of generation, such as code generation \citep{vq-llm} and action generation \citep{vq-vla}. There are also works using VQ to extract a hierarchical concept in LLM \citep{conceptlm}. Previous work has shown great compression and reconstruction capability of VQ for continuous vectors. However, prior work has rarely explored the use of VQ for representing text features in LLMs or for compressing KV caches. Our VQKV leverages VQ for context representation and demonstrates strong empirical effectiveness.

\section{Compressing KV with Vector Quantization}
\begin{figure*}
    \centering
    \includegraphics[width=\textwidth]{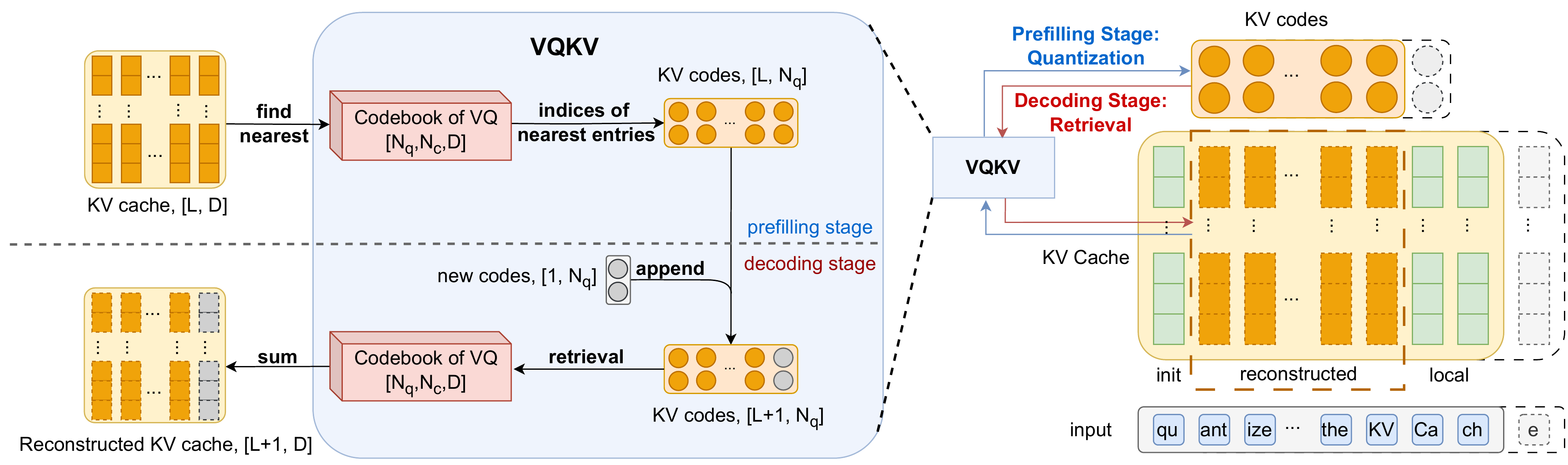}
    \caption{Overview of our VQKV. The left part shows the detailed process of our VQKV on prefilling stage and decoding stage. The right part shows the overview of our VQKV.}
    \label{fig:method}
\end{figure*}

In this section, we introduce how VQKV employs vector quantization to compress the KV cache to a high compression ratio while enabling high-fidelity reconstruction. We first describe how VQ is learned for compressing the KV cache, followed by a detailed discussion of how VQKV is integrated into the inference process, on both prefilling and decoding stages. Figure \ref{fig:method} illustrates the overall workflow of VQKV, where the \textbf{left} side depicts the compression–reconstruction process of the KV cache, and the \textbf{right} side presents how VQKV is integrated into the prefill and decoding stages. During \textbf{prefill}, VQKV compresses each KV cache vector by mapping it to the nearest entries in multiple codebooks and storing only the corresponding indices as KV codes. During \textbf{decoding}, VQKV compresses the cache of each new token and updates the stored KV codes while maintaining a \textit{local} sliding window by discarding the oldest entries.
When generating new tokens, VQKV performs an on-demand reconstruction of the KV cache from the codebooks.

\subsection{Learning High-fidelity VQ Codebooks}
We use Residual Simple Vector Quantization (RSimVQ) to compress each KV vector. Combining Residual VQ \citep{soundstream} with SimVQ \citep{simvq}, RSimVQ consists of multiple residual-connected codebooks and has an extra projection matrix on every codebook. On compressing, each codebook can represent an independent subspace of the original cache by an index of the nearest codebook entry. While retrieving, RSimVQ sums up all the codebook entries retrieved from codebooks to reconstruct the original cache.

Formally, let $x \in \mathbb{R}^D$ be one of the key/value vectors from KV Cache and $\hat{x}$ be the reconstruction, $Q_i$ be the codebook with entries $q_1,..., q_{S}$ where $S$ is the codebook size. For each codebook, RSimVQ first finds the nearest entry $q_z$ and records corresponding index $z$.
\begin{equation}
    \hat{x}=\operatorname{argmin}_{q\in\{q_1,...,q_{S}\}}\|x-Wq\| \triangleq Wq_z
\end{equation}
where $W$ is the projection matrix of each codebook and $z$ be the selected index. Then, RSimVQ iteratively sends the residual part $(x-\hat{x})$ to $Q_{i+1}$ for next step quantization.
\begin{equation}
    x\leftarrow x-\hat{x},\quad z_i\leftarrow z
\end{equation}

After $N$ iterations, RSimVQ could finally map the original cache into a few set of integer codes $z_1,...,z_{N}$. VQKV only stores codebooks and these quantized codes in the memory instead of the original cache vectors, leading the memory usage of the original KV cache from $[L,D]$ floats (ignoring layer and head dimension here for simplicity) into $[L, N]$ integers, which would reduce a lot of memory footprint when the KV cache vectors are massive. When reconstructing, RSimVQ simply looks up entries according to these codes and sums these entries up.
\begin{equation}
    \hat{x} = \sum_i^{N}Q_i[z_i]
\end{equation}

We prefetch the KV cache from around 10M tokens to train all these codebooks and two sets of codebooks are trained separately for key and value cache. During training, RSimVQ uses a stop-gradient operation $\operatorname{sg}(\cdot)$ to keep gradient flowing after discretization, along with the training loss as follows:
\begin{equation}
    \mathcal{L}=\|x-\hat{x}\|^2+\beta\|q_z-\operatorname{sg}(x)\|^2+\gamma\|x-\operatorname{sg}(q_z)\|
\end{equation}

\subsection{Prefilling Stage}
The codebooks used in VQKV are trained independently and can be directly integrated without modifying the original model parameters during inference, making VQKV a training-free approach. Therefore, the model’s performance is only minimally affected, as long as the compression algorithm can reconstruct the KV cache with sufficient fidelity.

We preserve both an \textit{initial} segment of the KV cache of length $L_{init}$ and the most recent segment of length $L_{local}$ from compression (seen in the right part of Figure \ref{fig:method}). For simplicity, the following discussion considers only the \textit{intermediate} KV cache to be compressed, unless stated otherwise.

As discussed above, we use two sets of codebooks to compress \textit{intermediate} cache $K^{m}, V^{m}$ into KV codes $K^{c}, V^{c}$ correspondingly. Take the key cache compression as an example. Let the codebook of the key RSimVQ be $Q^k_i$ and the codebook consists of $S_k$ indenpendent entries $q_1,q_2,...,q_{S_k}$. When prefilling, every cache vector $k\in K^{m}$ finds a nearest entry in the first codebook $Q^k_1$ and gets a correlated index $q^k_1$.
\begin{equation}
    q^k_1=\text{argmin}_{q\in Q^k_1}\|k-W^k_1q\|
\end{equation}

Then, the residual part $(k-q^k_1)$ is sent to codebook $Q^k_2$ for next step quantization.
\begin{equation}
    k \leftarrow (k-q^k_1)
\end{equation}

After $N^k$ times of iterations, the original vector $k$ can be represented as $N^k$ numbers of integers $(q^k_1, q^k_2, ..., q^k_{N^k})$, each of which is not greater than $S^k$. As a result, vectors that originally had a dimensionality of $\mathbb{R}^{D^k}$ can now be represented using only $N^k$ integers, with each integer occupying only $\log S^k$ bits.

The whole compression process can be expressed iteratively as follows.
\begin{equation}
    \begin{aligned}
        q^k_i=\text{argmin}_{q\in Q^k_i}&\|k_i-W^k_iq\| \\
        k_1=k,\quad k_{i+1}=k_i-q_i^k&,\quad i=1,...,N^k
    \end{aligned}
\end{equation}
\begin{equation}
    q^k=\mathcal{VQ}^k(k)=(q^k_1,q^k_2,...,q^k_{N^k})
\end{equation}

This procedure is applied to every key/value vector, compressing the original floating-point representations of size $K^m \in \mathbb{R}^{L\times D^k}$ into a small number of integer indices $K^{c}\in\mathbb{N}^{L\times N^k}$, significantly reducing the memory cost of KV Cache storage. After compressing, we use the \textbf{original} KV for forward propagation on prefilling stage.
\begin{equation}
    O=\operatorname{flash\_attention}(Q,K,V)
\end{equation}

\subsection{Decoding Stage}
In the decoding stage, we first compress the cache out of the \textit{local} segment and concatenate them to the previous compressed KV codes $K^c$. Then VQKV reconstructs the \textit{intermediate} KV cache by selecting the entries from each codebook according to the stored indices. We continue to use the key cache as an illustrative example. For each stored key indices $q_k \in K^c$, the reconstructed key vector $\hat{k}$ is calcuated by summarizing all entries in the codebooks.
\begin{equation}
    \hat{k} = \mathcal{VQ}^k[q^k] = \sum_{i=1}^{N^k}(Q^k_i[q^k_i])
\end{equation}

With FlashAttention \citep{flashattention,flashattention2}, it is unnecessary to reconstruct the full KV cache for the entire sequence at each decoding step. Instead, when reconstructing the cache, we only rebuild the portion required for the current decoding step, aligned with the FlashAttention computation flow, thereby further reducing the memory footprint of VQKV.
\begin{equation}
    \hat{k}_m = \mathcal{VQ}^k[q^k_m],\quad \hat{v}_m = \mathcal{VQ}^v[q^v_m]
\end{equation}
\begin{equation}
    o_m = \operatorname{flash\_attention}(q_{t+1}, \hat{k}_m, \hat{v}_m)
\end{equation}

A customized FlashAttention kernel is designed and implemented for VQKV, further improving the memory efficiency of VQKV and carefully balancing the additional computational overhead introduced by vector quantization. By integrating FlashAttention, the decoding process no longer requires reconstructing the entire sequence back to full precision. Instead, only the portions of the KV cache needed for the current attention computation are reconstructed, which is sufficient to complete the attention operation. This design further reduces memory consumption, while the fusion of vector quantization operators also mitigates the additional computational overhead.

To better improve the efficiency of the quantization process, we optimize the algorithm for computing the nearest distances within codebooks. Repeating the compression and concatenation process at each decoding step introduces significant computational overhead. By adopting a block-wise computation strategy, we effectively reduce the peak memory consumption during distance calculations. Moreover, since the residual structure in RSimVQ inherently lacks parallelism, we enhance parallel efficiency in the quantization process by performing batched quantization computations based on the current $L_{local}$. Specifically, instead of quantizing KV cache one by one during LLM decoding, we compute the quantization process on the entire cache segment $K^l$ and $V^l$ in a single operation. In this way, the compression step of VQ is performed only once every $L_{local}$ steps, which would otherwise occur at every decoding step, thereby substantially improving efficiency.

\subsection{Compression Ratio}
The total compression ratio of VQKV is associated with the number and the size of codebooks. For each vector in the KV cache, the original representation consists of $D$ 16-bit floating-point values, where $D$ refers to head dimensions. After compression, each vector is represented using only $N$ integers, where each integer indexes a codebook of size $S$. Consequently, the compressed representation requires only $N \times \log S$ bits. For KV cache with dimensions of $D^k$ and $D^v$, our VQKV achieves a compression ratio $r$ such that
\begin{equation}
    r=\Big(1-\frac{N^k\log S^k+N^v\log S^v}{16(D^k+D^v)}\Big)\times100\%
    \label{compressionratio}
\end{equation}

\section{Experiment}
In this section, we evaluate VQKV across different models and downstream tasks. We first compare our method with three feature dimension reduction approaches under the same compression ratio setting: ASVD \citep{asvd}, Palu \citep{palu} and KIVI \citep{kivi}. In addition, since SnapKV \citep{snapkv} is a strong baseline, we also include SnapKV in our comparison, even though it is not a feature dimension reduction method. Then, we evaluate our memory efficiency by testing the maximum generation length against the full-cache model under identical hardware conditions. Finally, ablation studies on both the number and size of codebooks provide insights into the factors affecting VQKV.

\subsection{Setup}
\label{training_detail}
We conduct all our experiments on LLaMA3.1-8B \citep{llama3.1-8b} and LLaMA3.2-3B \citep{llama3.2-3b}. For both models, we set the length of initial token $L_{init}$ to 4 and the length of local tokens $L_{local}$ to 1024. We sample 0.1\% data of OpenWebText \citep{openwebtext} for RSimVQ training and every RSimVQ is trained with learning rate 0.001 and batch size 65536. The size of codebooks vary from cache type and models. For LLaMA3.1-8B \citep{llama3.1-8b}, we use the codebook number $(N^k,N^v)=(56, 16)$ with the codebook size $(S^k, S^v)=(1024, 512)$, while for LLaMA3.2-3B \citep{llama3.2-3b}, we use a codebook number $(N^k,N^v)=(56, 10)$ and set all codebook sizes to $(S^k, S^v)=(1024, 65536)$. By Equation \ref{compressionratio}, our method achieves a compression ratio of 82.8\% on LLaMA3.1-8B and 82.4\% on LLaMA3.2-3B.

\begin{table*}[htbp]
    \centering
    \setlength{\tabcolsep}{1.9pt}
    \small
    \caption{Results of LLaMA3.1-8B and LLaMA3.2-3B on LongBench . The \textbf{Ratio} means discarding ratios. Our VQKV achieves the closest performance to the full cache models on comparable compression ratio against other methods.}
    \resizebox{\textwidth}{!}{
    \begin{tabular}{@{}l l*{18}{c}@{}}
        \toprule
        & & \multirow{2}{*}{\textbf{Ratio}}
          & \multicolumn{3}{c}{\textbf{Single-Doc}}
          & \multicolumn{3}{c}{\textbf{Multi-Doc}}
          & \multicolumn{3}{c}{\textbf{Summary}}
          & \multicolumn{3}{c}{\textbf{Few-shot}}
          & \multicolumn{2}{c}{\textbf{Synthetic}}
          & \multicolumn{2}{c}{\textbf{Code}}
          & \multirow{2}{*}{\textbf{Avg.}} \\
        \cmidrule(lr){4-6} \cmidrule(lr){7-9} \cmidrule(lr){10-12} \cmidrule(lr){13-15} \cmidrule(lr){16-17} \cmidrule(lr){18-19}
        \multicolumn{2}{c}{} &  & \textbf{NQ} & \textbf{Qsp} & \textbf{MF}
        & \textbf{HQ} & \textbf{WQ} & \textbf{Msq}
        & \textbf{GR} & \textbf{QS} & \textbf{MN}
        & \textbf{TR} & \textbf{TQ} & \textbf{SS}
        & \textbf{PC} & \textbf{PR} 
        & \textbf{LCC} & \textbf{Re-P} \\
        \midrule
        \multirow{5}{*}
        &\textit{LLaMA3.1-8B} & 0 &\gray{13.0} &\gray{20.4} &\gray{32.1} &\gray{12.0} &\gray{14.0} &\gray{8.7} &\gray{29.7} &\gray{25.2} &\gray{1.0} &\gray{73.5} &\gray{91.0} &\gray{47.2} &\gray{0.8} &\gray{26.8} &\gray{72.0} &\gray{69.2} &\gray{33.5}\\
        & + SnapKV            & - & 12.7 & 19.5 & \textbf{31.2} & 11.5 & \underline{13.9} & \textbf{8.9} & \underline{23.7} & \textbf{24.1} & 0.9  & 70.0 & \textbf{91.3} & \textbf{46.5} & 0.9 & \underline{27.8} & \underline{71.7} & \underline{68.5} & \underline{31.7} \\
        & + KIVI              & 75.0\% & \textbf{19.2} & \textbf{20.8} & \textbf{31.2} & \textbf{15.2} & \textbf{17.7} & \underline{8.2} & 17.6 & 12.1 & 3.9  & \underline{72.5} & 89.6 & 33.9 & \underline{2.2} & 15.9 & 67.6 & 64.4 & 30.7 \\
        & + ASVD              & 80.0\% & 4.6  & 9.9  & 16.1 & 9.7  & 7.4  & 5.2 & 9.0  & 16.6 & \textbf{12.8} & 60.0 & 78.7 & 33.7 & \textbf{2.8} & 4.9  & 30.4 & 36.9 & 21.2 \\
        & + Palu              & 80.0\% & 6.4  & 16.6 & 23.1 & 9.7  & 12.3 & 6.6 & 16.5 & 21.5 & \underline{10.7} & \underline{72.5} & 84.6 & 37.1 & 1.3 & 14.3 & 64.7 & 59.1 & 28.6 \\
        & + \textbf{VQKV(Ours)}& 82.8\% & \underline{13.9} & \underline{19.8} & 29.9 & \underline{11.8} & 13.8 & 8.1 & \textbf{26.0} & \underline{23.6} & 0.9  & \textbf{73.5} & \underline{90.5} & \underline{45.9} & 0.8 & \textbf{29.7} & \textbf{71.8} & \textbf{69.2} & \textbf{33.1} \\
        \addlinespace[2pt]
        \midrule
        \multirow{5}{*}
        &\textit{LLaMA3.2-3B} & 0 &\gray{10.3} &\gray{21.6} &\gray{35.5} &\gray{9.5} &\gray{12.8} &\gray{6.8} &\gray{30.1} &\gray{23.8} &\gray{28.3} &\gray{69.5} &\gray{87.2} &\gray{38.3} &\gray{0.0} &\gray{7.0} &\gray{70.0} &\gray{66.1} &\gray{32.3}\\
        & + SnapKV            & - & 10.5 & \underline{21.6} & \underline{34.1} & 9.4 & \underline{12.6} & \textbf{7.2} & 23.1 & \textbf{24.1} & \underline{27.1} & 65.0 & \underline{87.2} & \underline{38.4} & 0.0 & \underline{7.0} & 69.8 & \underline{65.1} & \underline{31.4} \\
        & + KIVI              & 75.0\% & 7.4  & \textbf{22.1} & 33.1 & \underline{9.8} & 12.0 & 5.2 & 17.4 & 14.2 & 16.0 & \textbf{69.5} & \textbf{88.3} & 30.3 & \underline{1.2} & \textbf{7.4} & \underline{70.0} & \textbf{65.6} & 29.3 \\
        & + ASVD              & 80.0\% & 0.8  & 10.7 & 8.8  & 5.4 & 5.2  & 2.7 & 8.7  & 8.6  & 11.1 & 31.5 & 49.1 & 17.7 & \textbf{3.3} & 3.5 & 35.4 & 36.6 & 14.9 \\
        & + Palu              & 80.0\% & \textbf{12.2} & 16.7 & 27.7 & \textbf{9.9} & 11.6 & \underline{6.6} & \underline{24.2} & 20.0 & 19.4 & 58.0 & 86.4 & 41.4 & 0.3 & 4.7 & 56.2 & 54.3 & 28.1 \\
        & + \textbf{VQKV(Ours)}& 82.4\% & \underline{10.7} & 20.6 & \textbf{36.7} & 9.5 & \textbf{14.5} & \underline{6.6} & \textbf{29.0} & \underline{23.1} & \textbf{27.5} & \textbf{69.5} & \underline{87.2} & \textbf{38.5} & 0.0 & \underline{7.0} & \textbf{70.6} & 64.4 & \textbf{32.2} \\
        \bottomrule
    \end{tabular}
    }
    \label{tab-mainresult}
\end{table*}
\begin{figure}[htbp]
    \centering
    \begin{subfigure}{0.32\linewidth}
        \includegraphics[width=\linewidth]{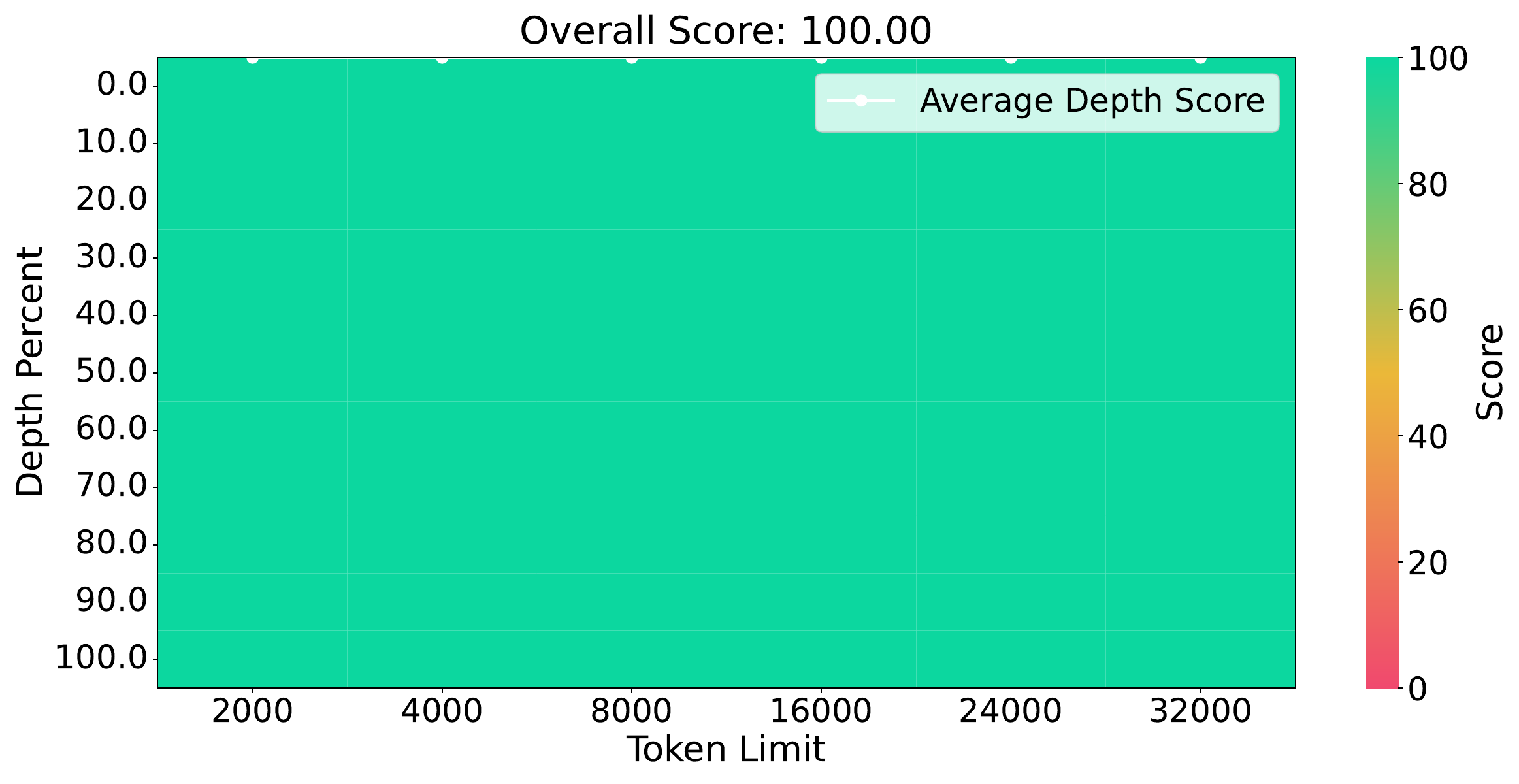}
        \caption{LLaMA3.1-8B}
    \end{subfigure}
    \begin{subfigure}{0.32\linewidth}
        \includegraphics[width=\linewidth]{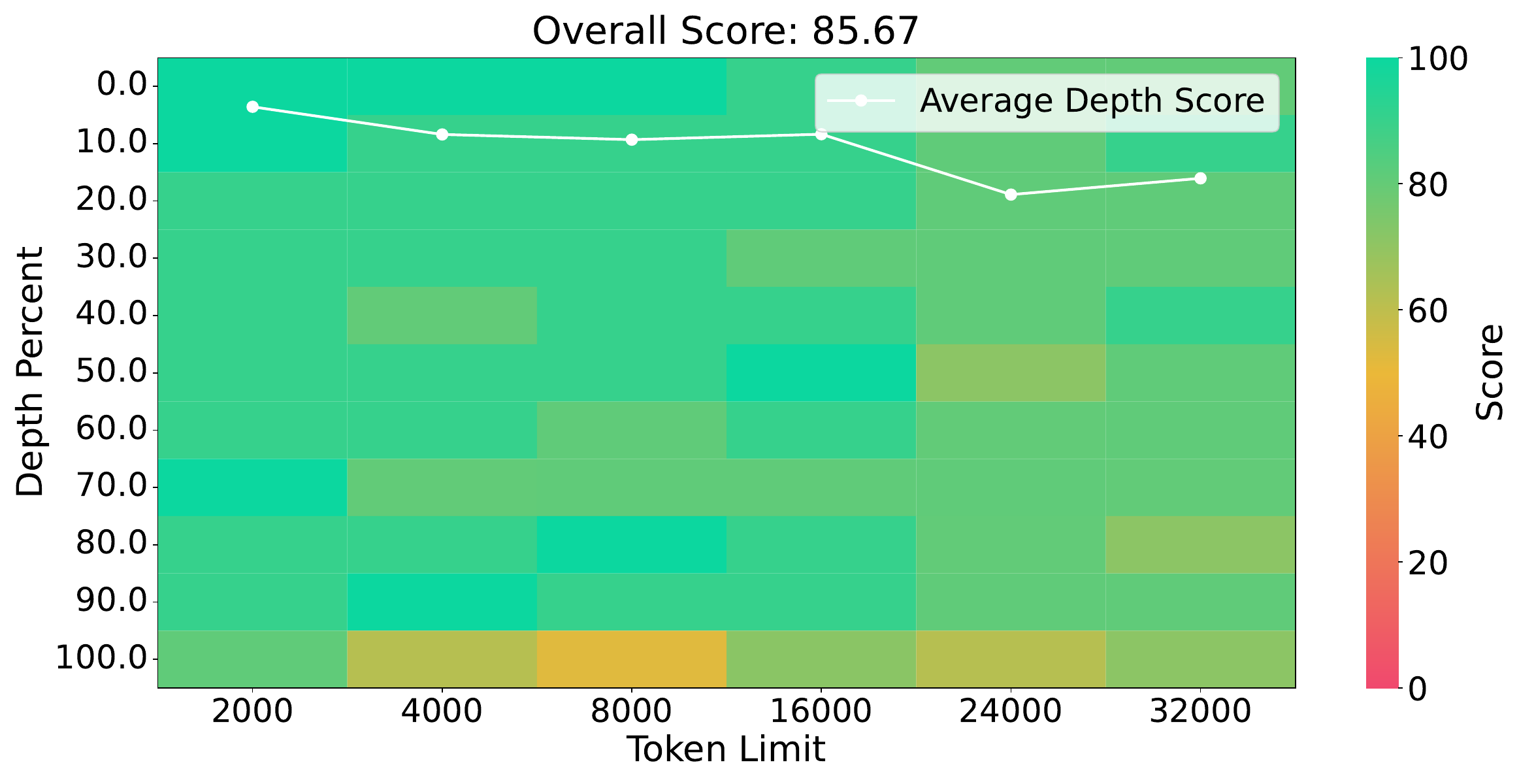}
        \caption{ASVD}
    \end{subfigure}
    \begin{subfigure}{0.32\linewidth}
        \includegraphics[width=\linewidth]{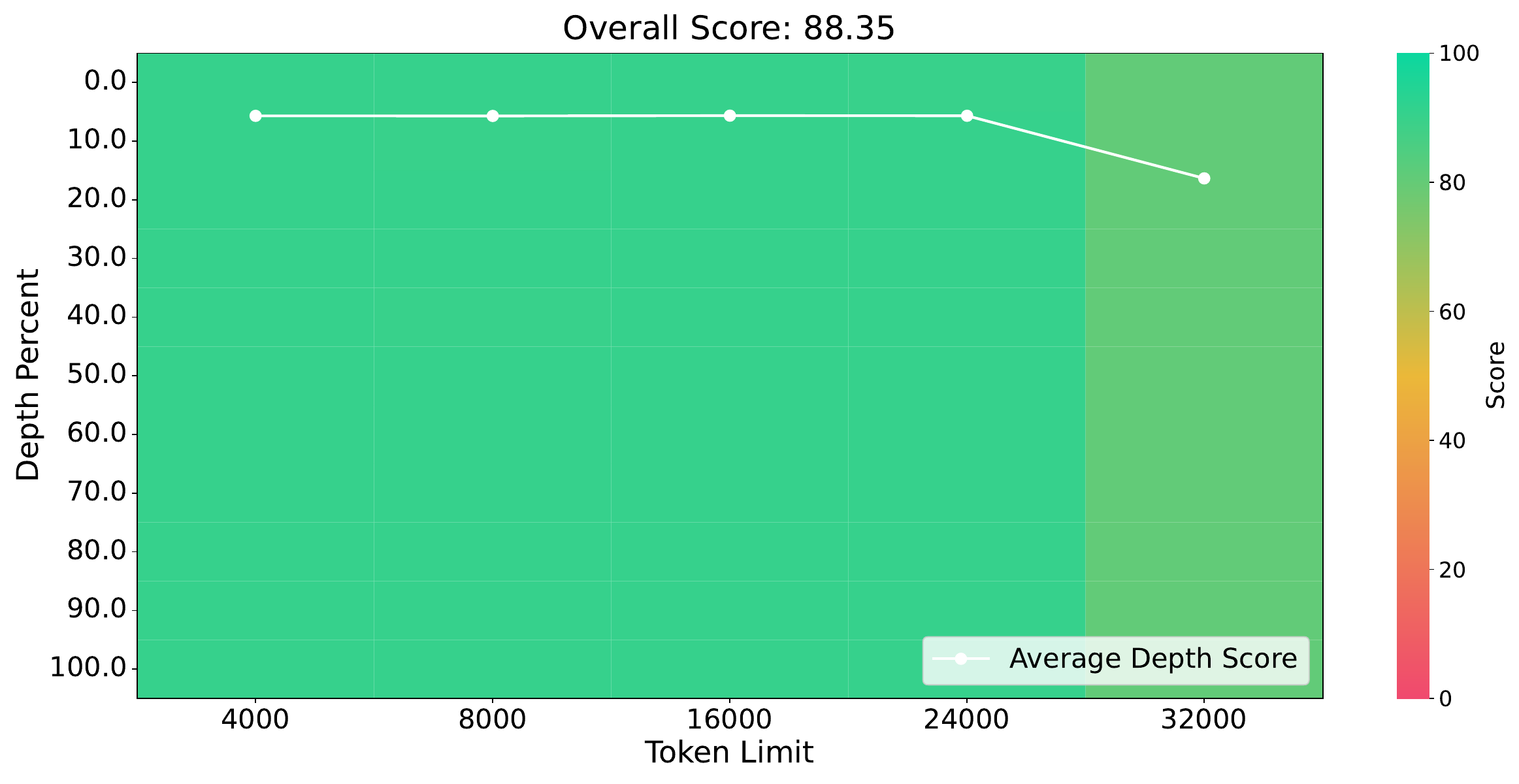}
        \caption{SnapKV}
    \end{subfigure}

    \vspace{0.2em} 
    \begin{subfigure}{0.32\linewidth}
        \includegraphics[width=\linewidth]{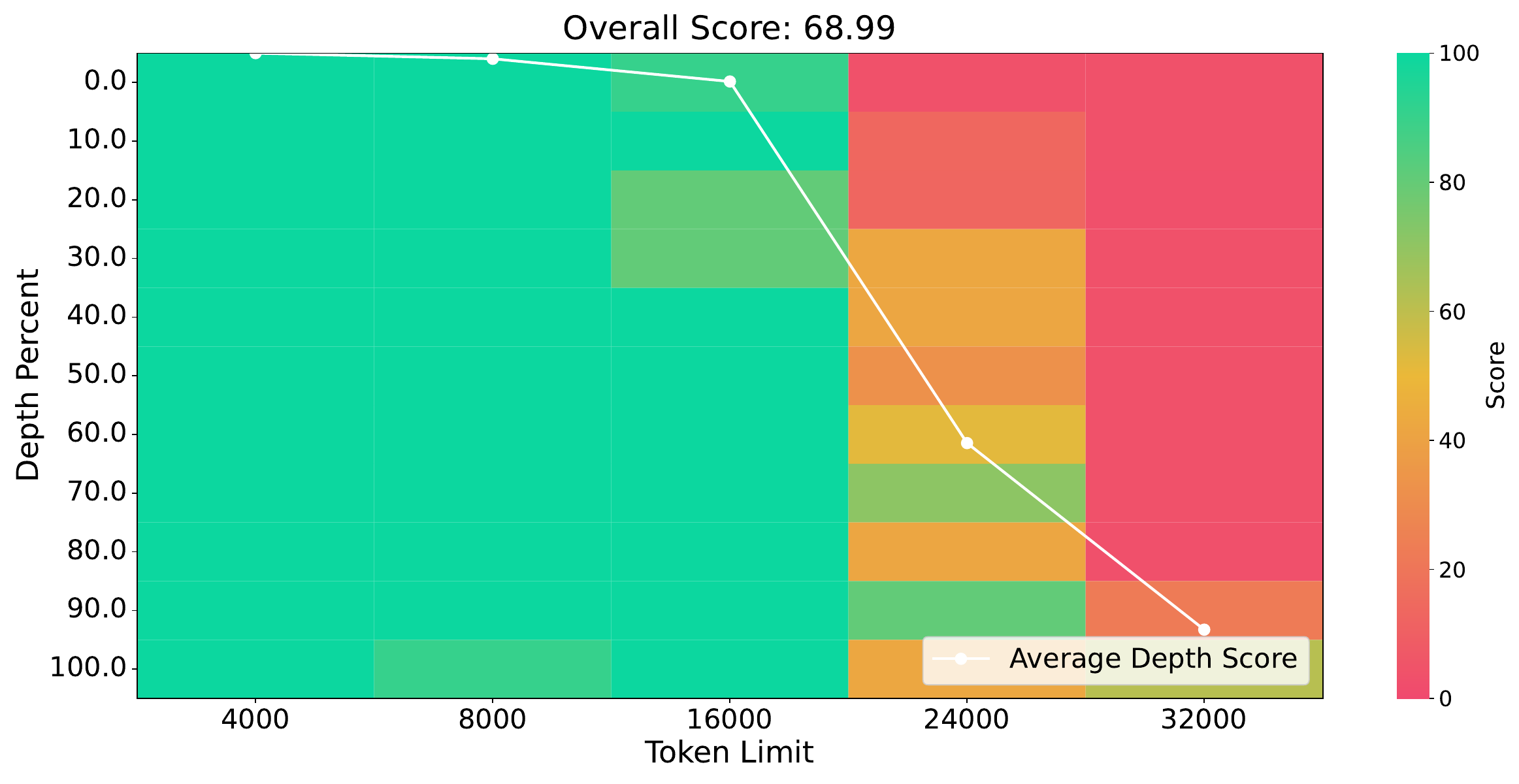}
        \caption{Palu}
    \end{subfigure}
    \begin{subfigure}{0.32\linewidth}
        \includegraphics[width=\linewidth]{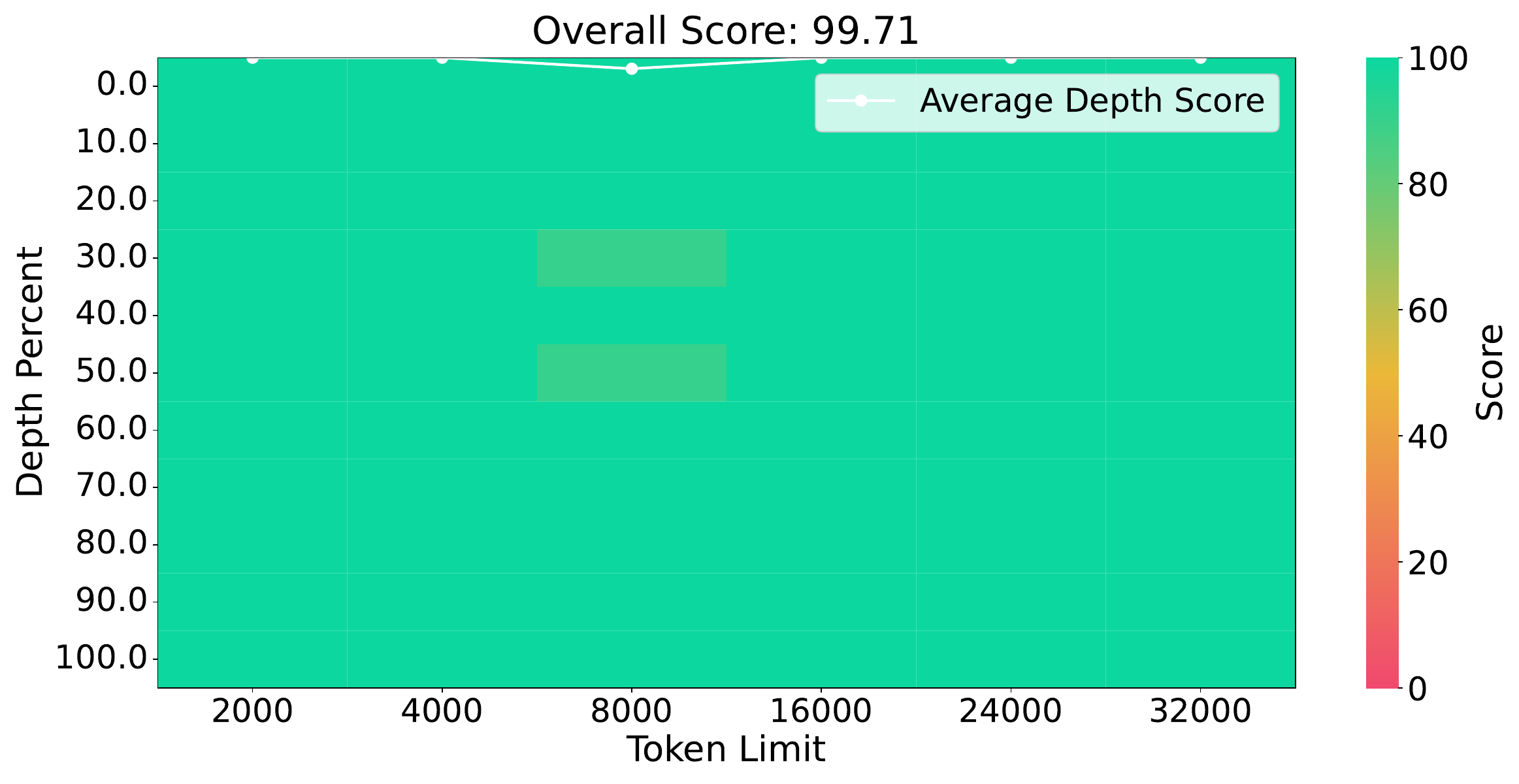}
        \caption{KIVI}
    \end{subfigure}
    \begin{subfigure}{0.32\linewidth}
        \includegraphics[width=\linewidth]{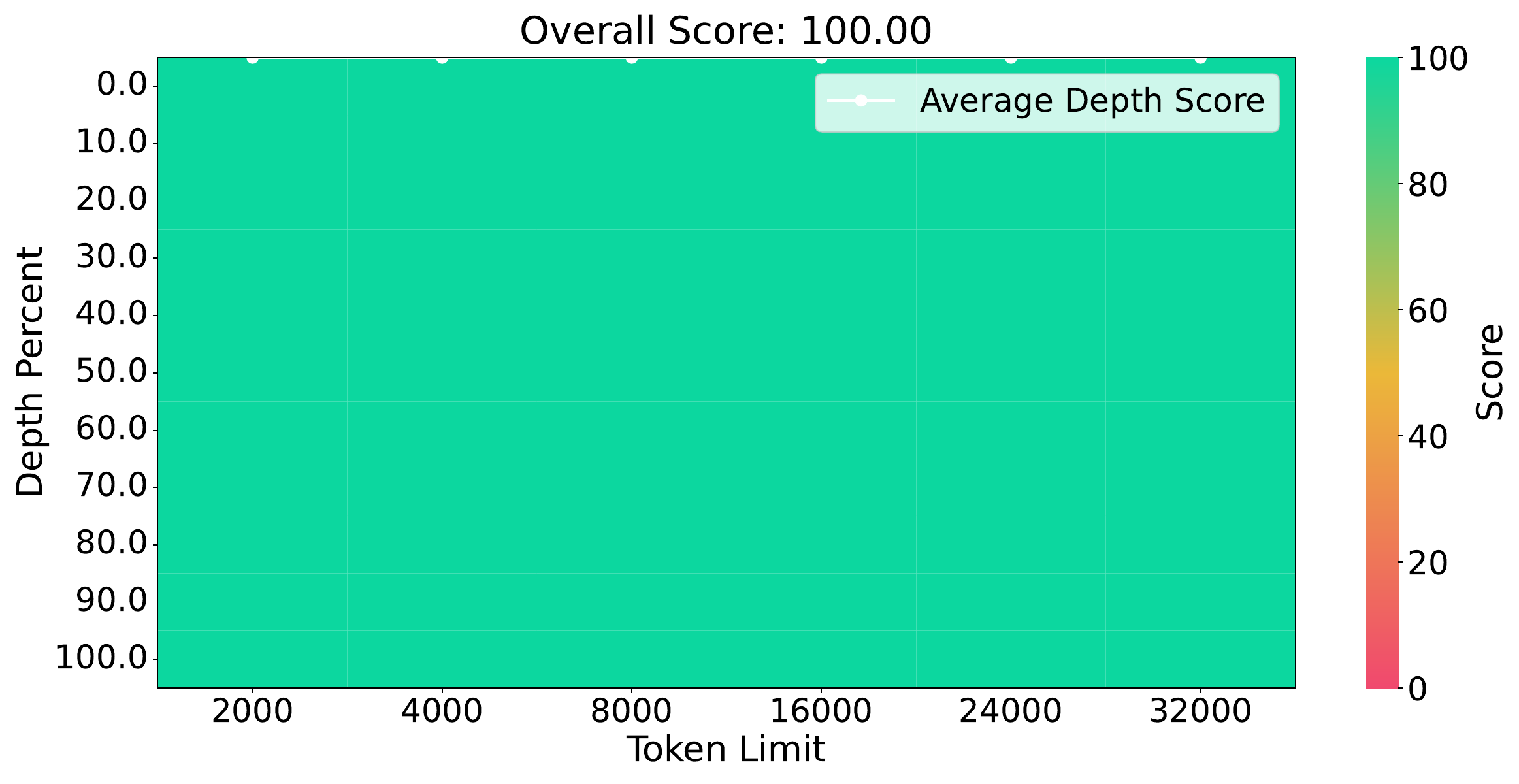}
        \caption{VQKV(ours)}
    \end{subfigure}

    \caption{Results of LLaMA3.1-8B on NIAH.}
    \label{fig:niah_llama3.1-8b}
\end{figure}
\begin{figure}[htbp]
    \centering
    \begin{subfigure}{0.32\linewidth}
        \includegraphics[width=\linewidth]{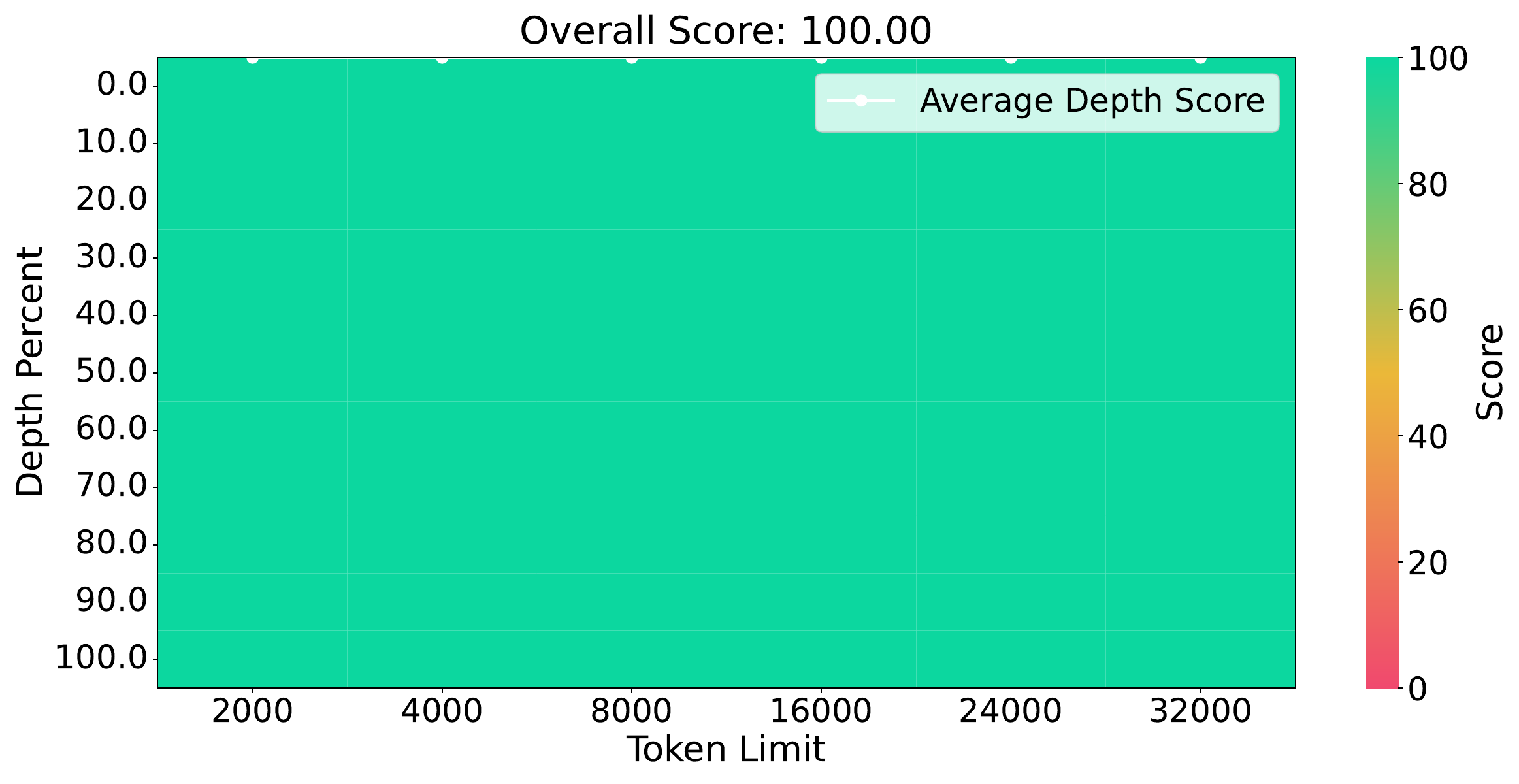}
        \caption{LLaMA3.2-3B}
    \end{subfigure}
    \begin{subfigure}{0.32\linewidth}
        \includegraphics[width=\linewidth]{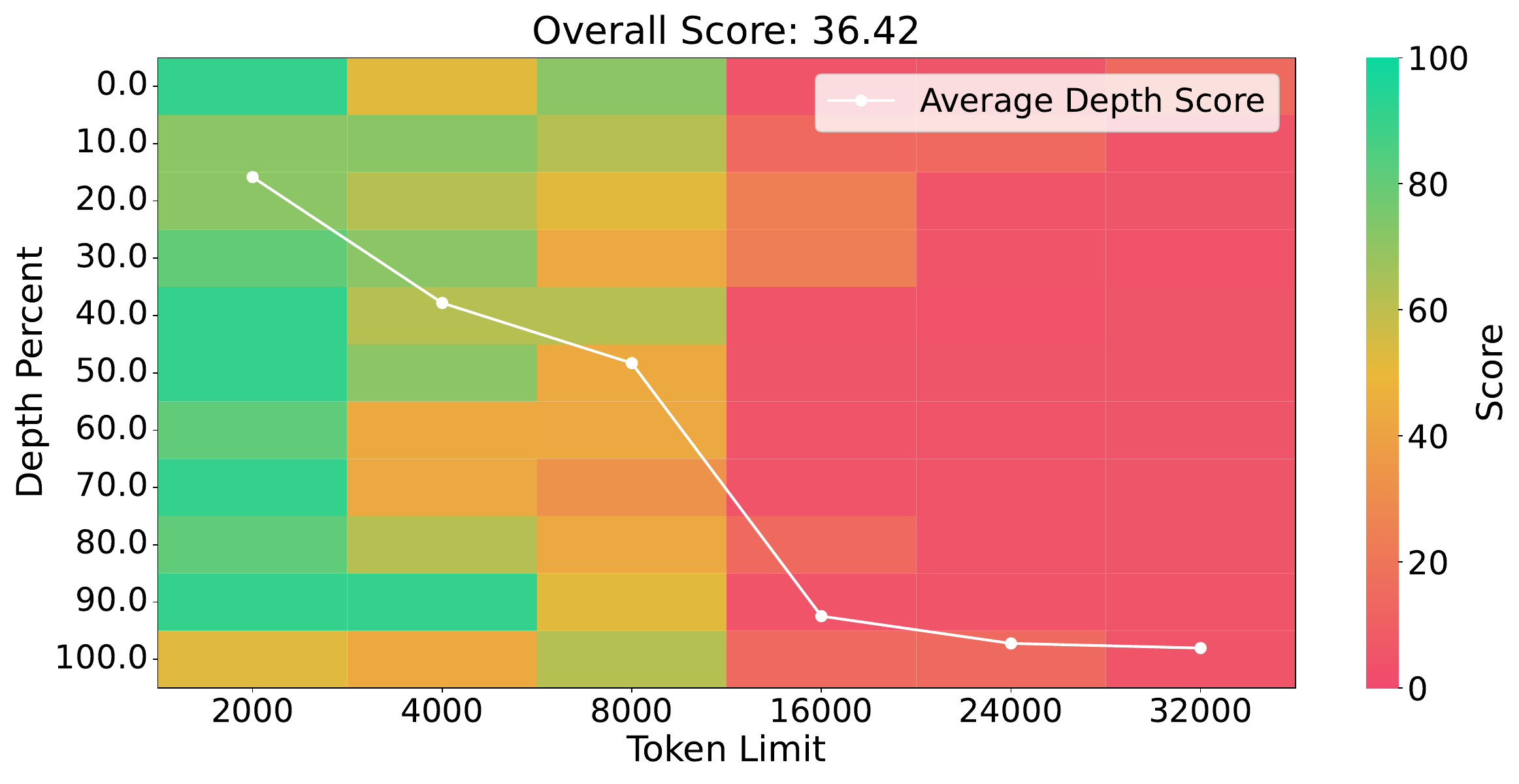}
        \caption{ASVD}
    \end{subfigure}
    \begin{subfigure}{0.32\linewidth}
        \includegraphics[width=\linewidth]{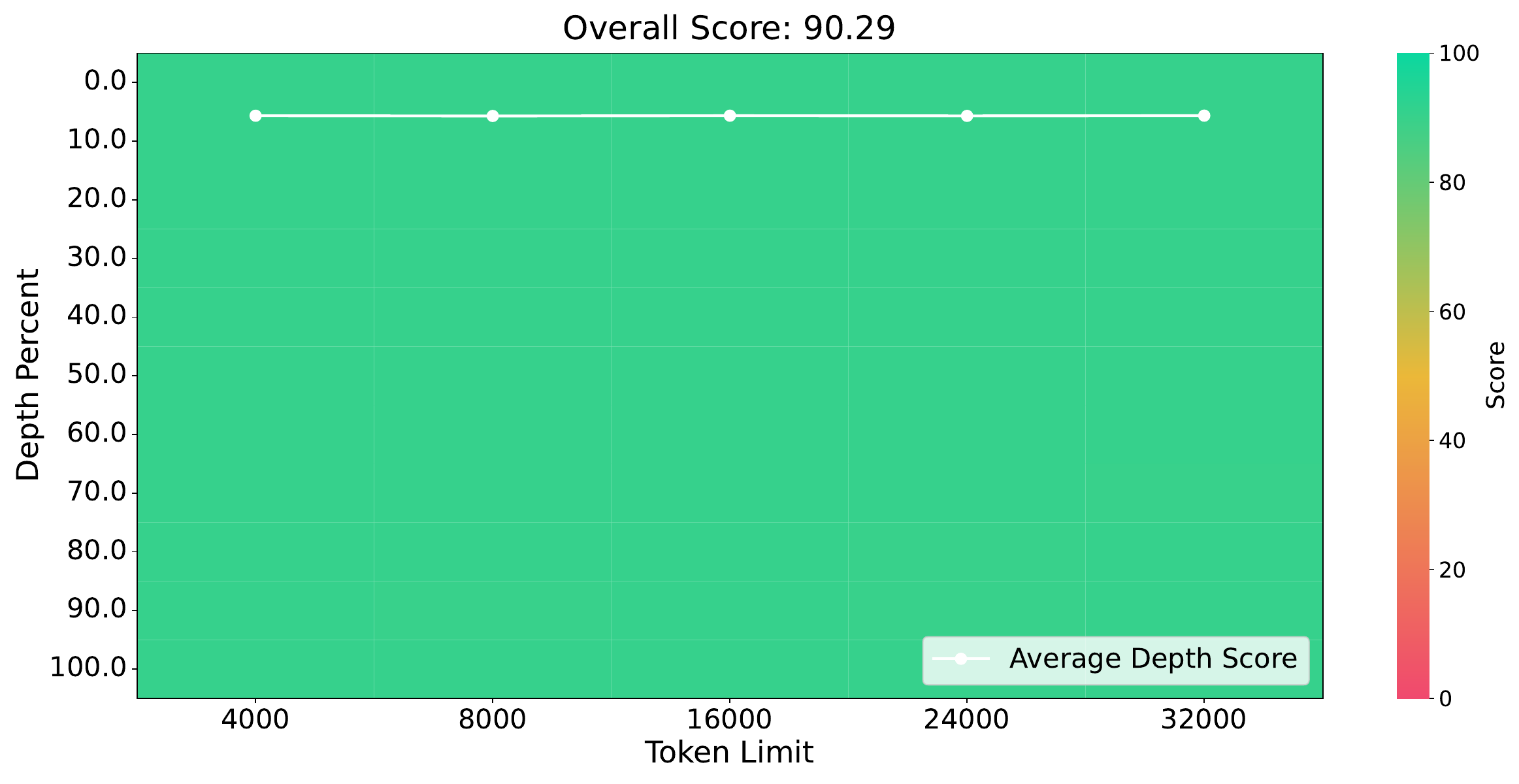}
        \caption{SnapKV}
    \end{subfigure}

    \vspace{1em} 
    \begin{subfigure}{0.32\linewidth}
        \includegraphics[width=\linewidth]{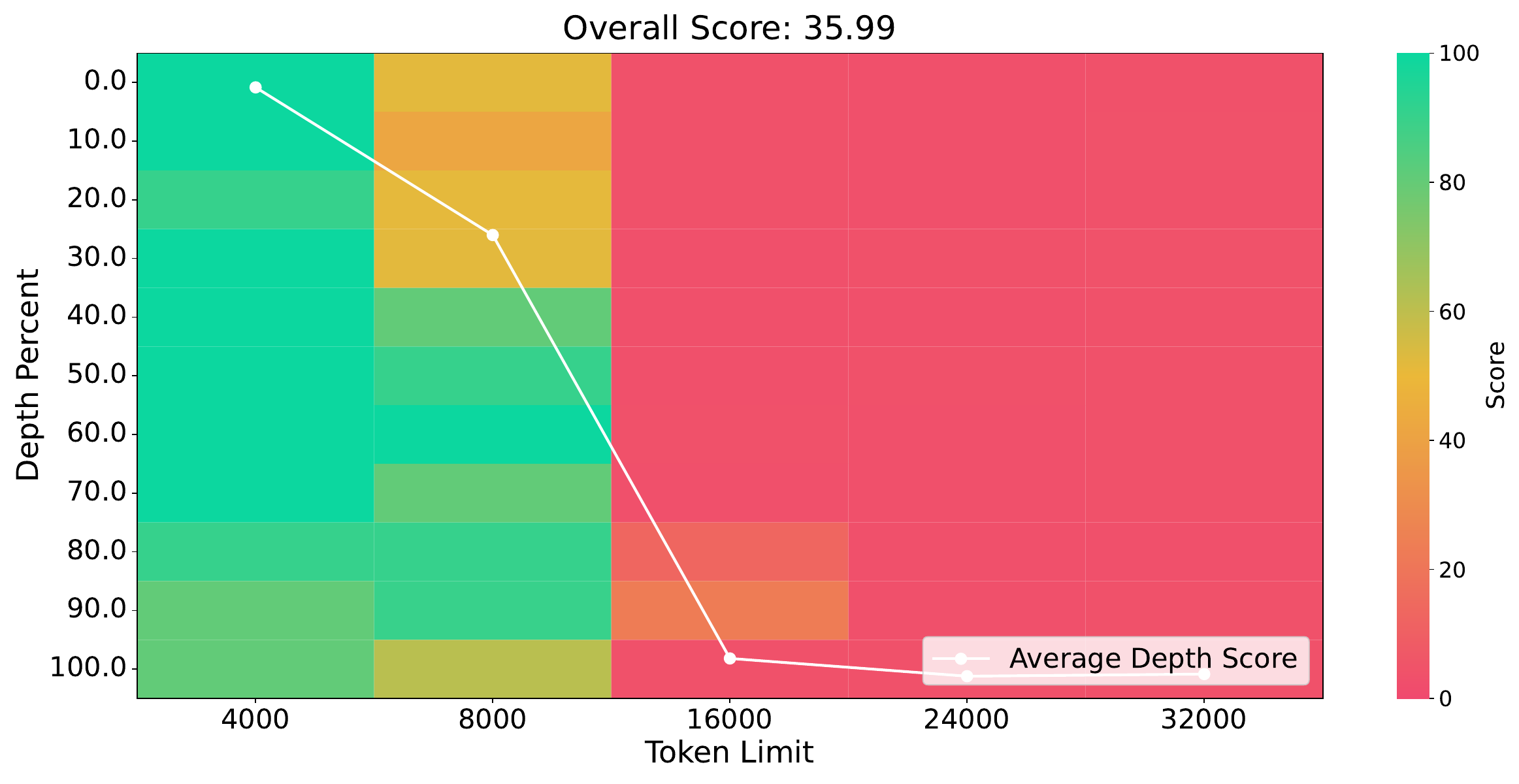}
        \caption{Palu}
    \end{subfigure}
    \begin{subfigure}{0.32\linewidth}
        \includegraphics[width=\linewidth]{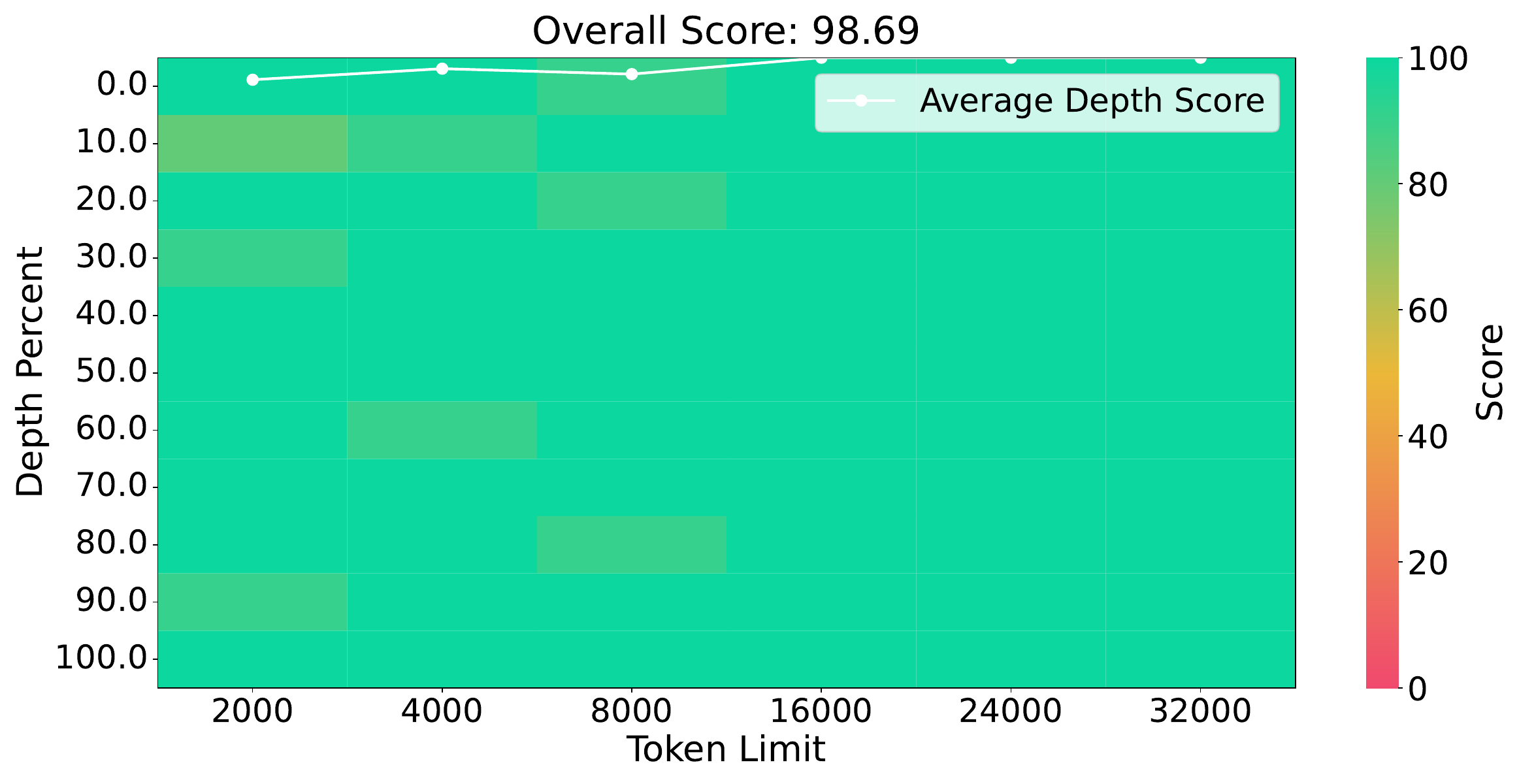}
        \caption{KIVI}
    \end{subfigure}
    \begin{subfigure}{0.32\linewidth}
        \includegraphics[width=\linewidth]{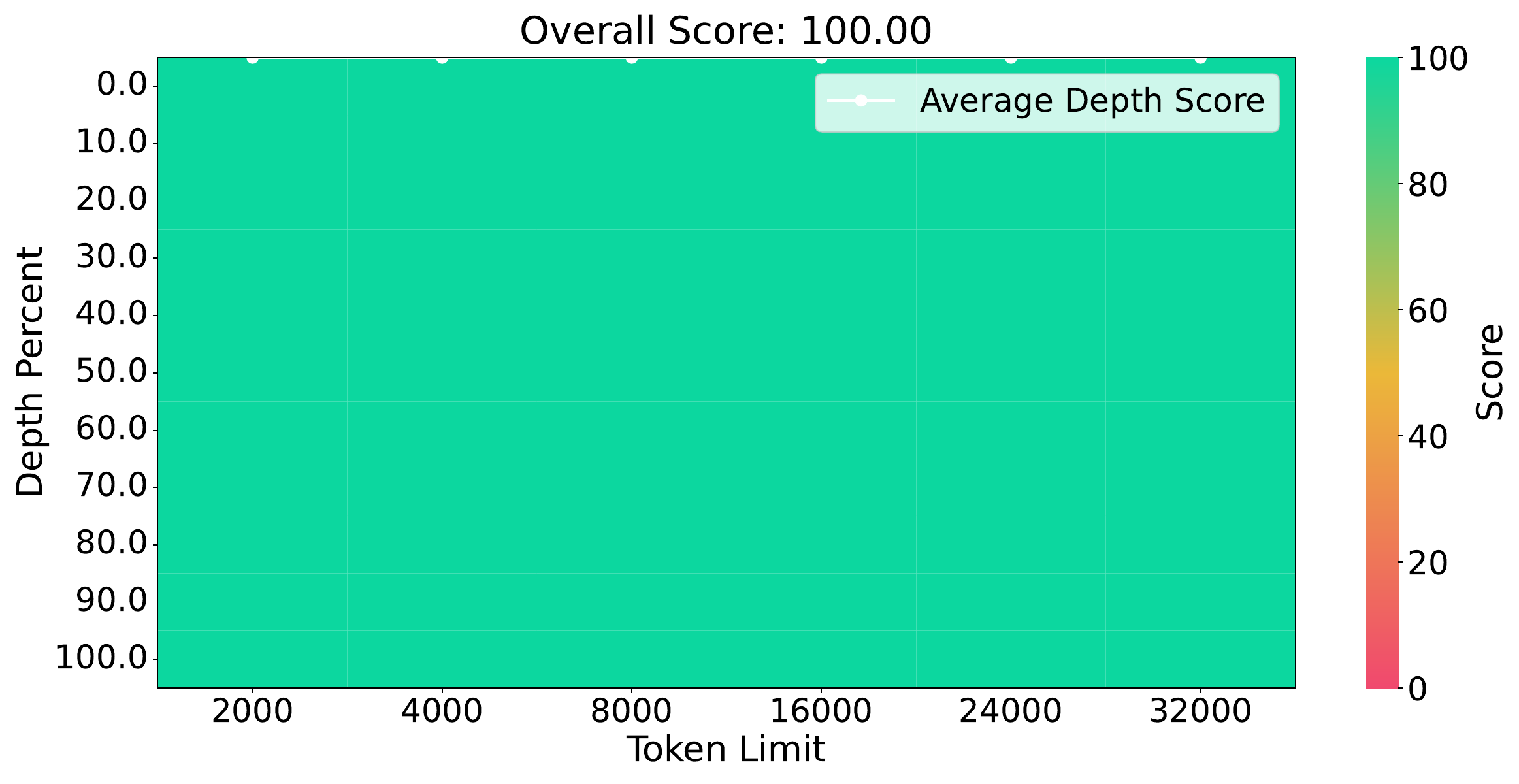}
        \caption{VQKV(ours)}
    \end{subfigure}

    \caption{Results of LLaMA3.2-3B on NIAH.}
    \label{fig:niah_llama3.2-3b}
\end{figure}
\begin{figure}[htbp]
    \centering
    
    \begin{subfigure}{0.24\linewidth}
        \includegraphics[width=\linewidth]{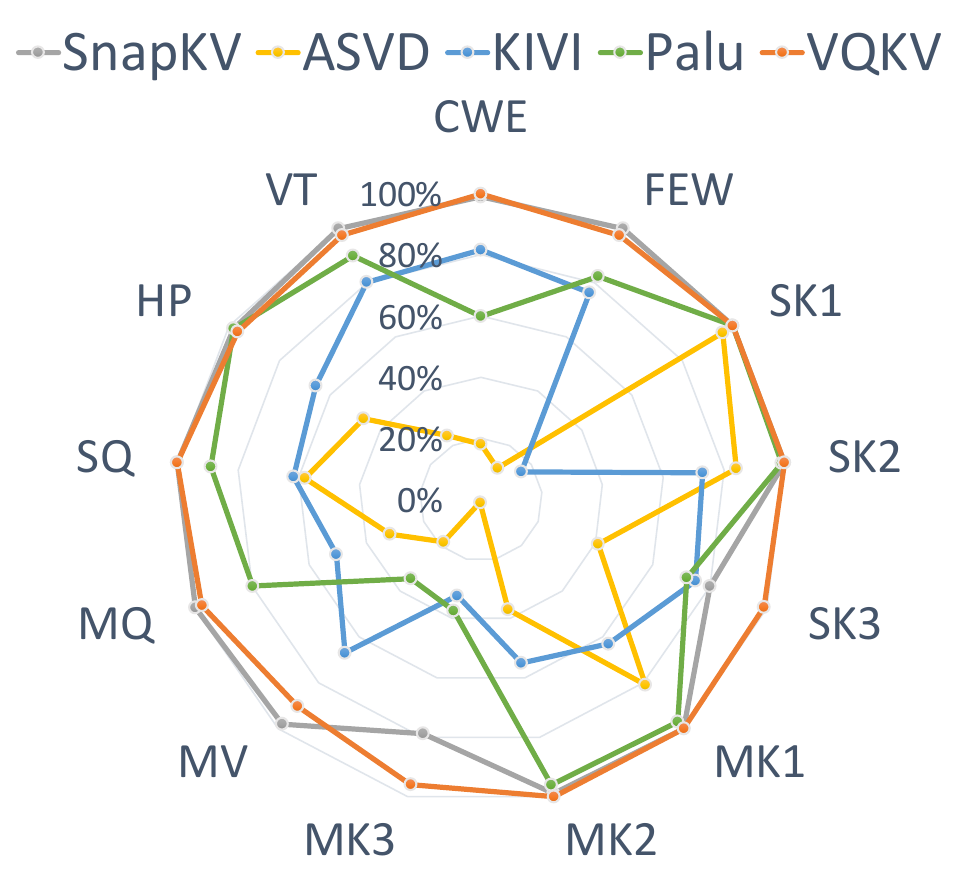}
        \caption{Results on RULER-4K.}
    \end{subfigure}
    \begin{subfigure}{0.24\linewidth}
        \includegraphics[width=\linewidth]{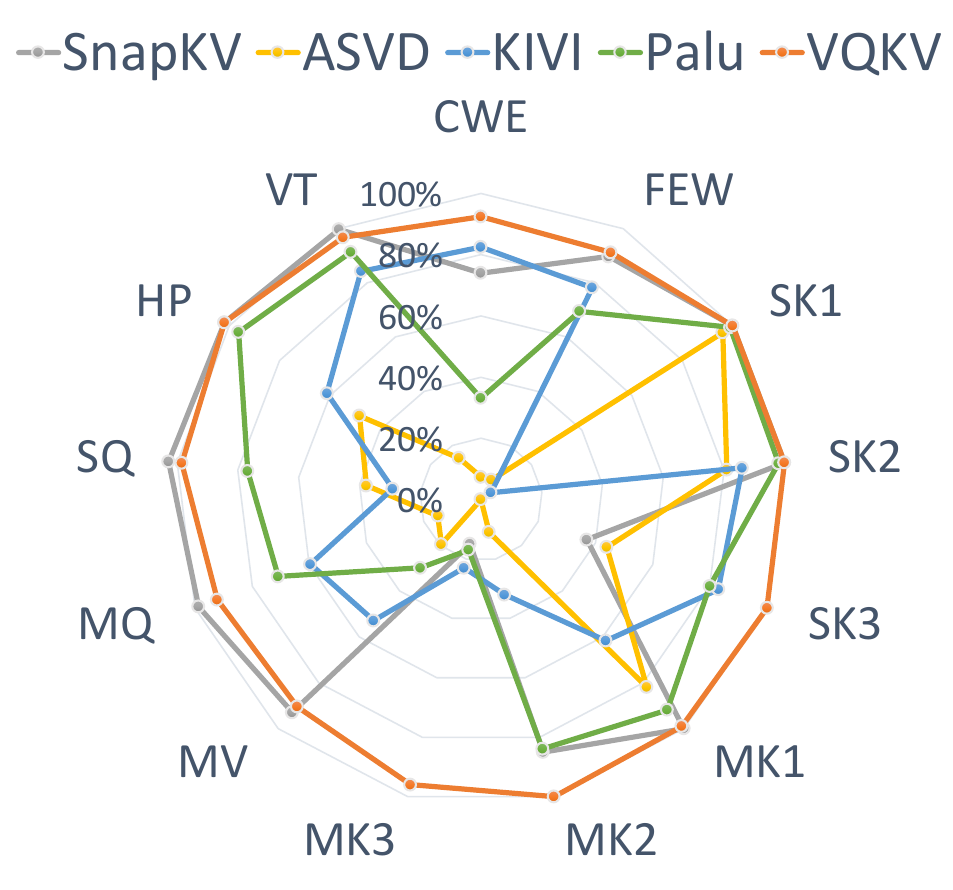}
        \caption{Results on RULER-8K.}
    \end{subfigure}
    \begin{subfigure}{0.24\linewidth}
        \includegraphics[width=\linewidth]{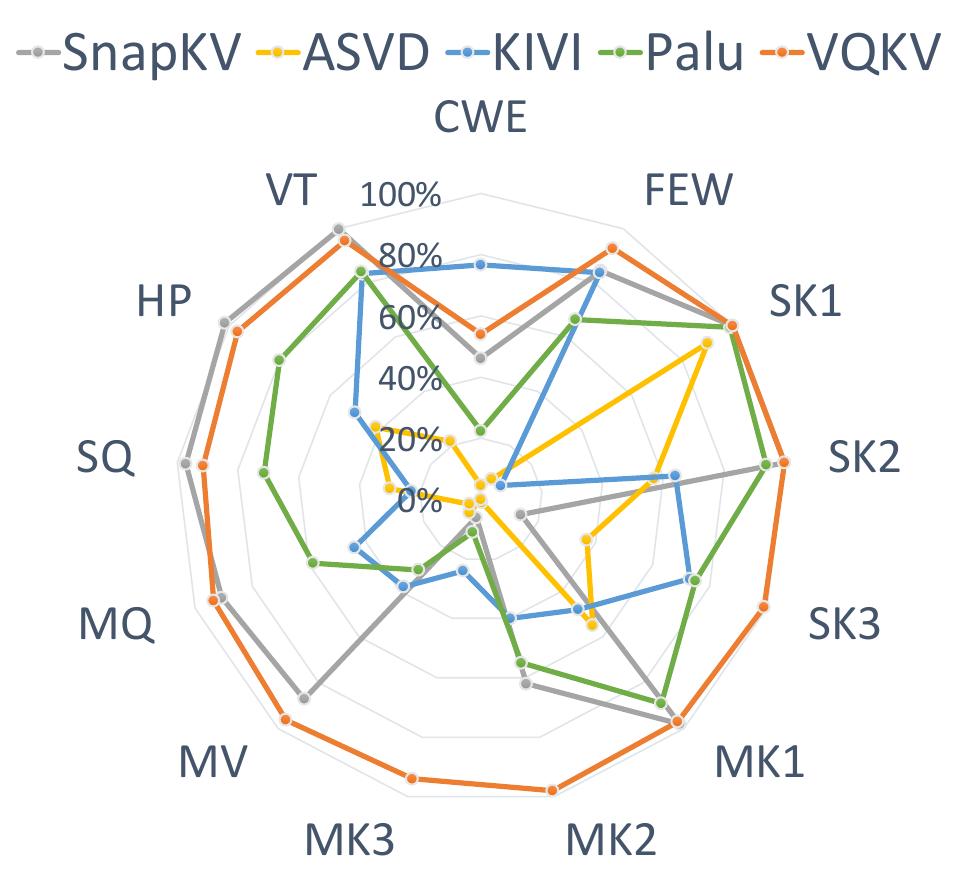}
        \caption{Results on RULER-16K.}
    \end{subfigure}
    \begin{subfigure}{0.24\linewidth}
        \includegraphics[width=\linewidth]{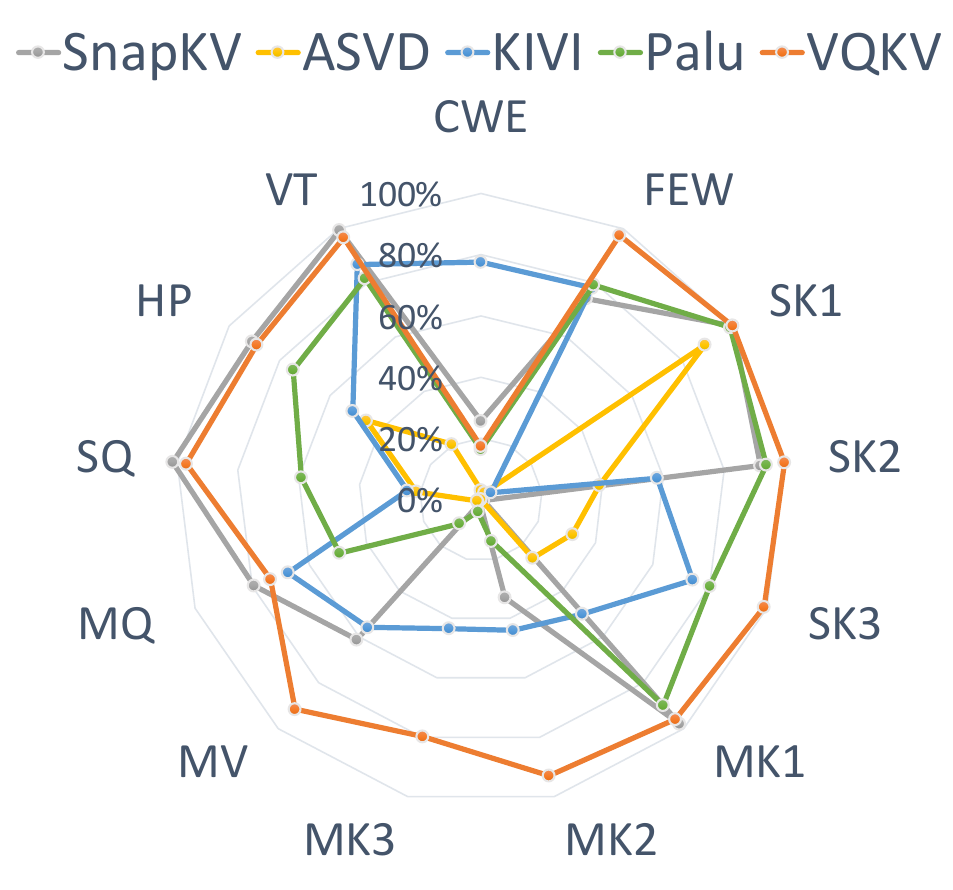}
        \caption{Results on RULER-32K.}
    \end{subfigure}

    \caption{Detailed results of LLaMA3.1-8B on RULER in different context length. The results of full cache model are taken as the 100\% reference.} 
    \label{fig:ruler1}
\end{figure}
\begin{figure}[htbp]
    \centering
    \begin{subfigure}{0.24\linewidth}
        \includegraphics[width=\linewidth]{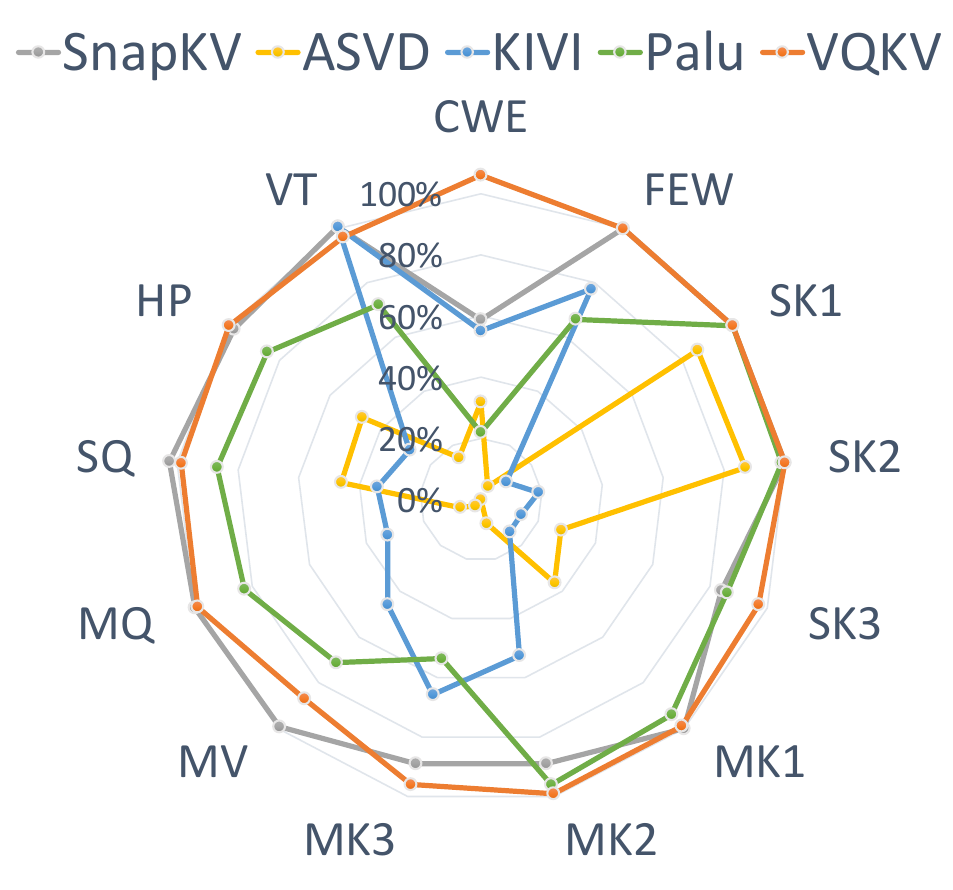}
        \caption{Results on RULER-4K.}
    \end{subfigure}
    \begin{subfigure}{0.24\linewidth}
        \includegraphics[width=\linewidth]{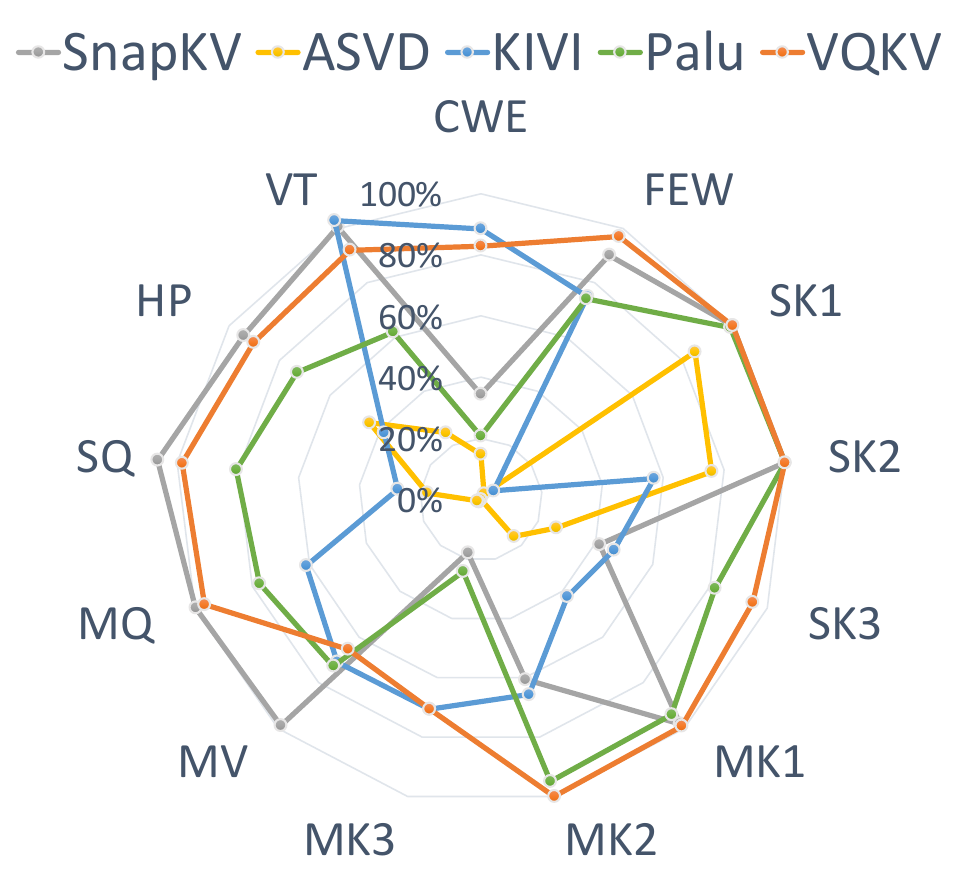}
        \caption{Results on RULER-8K.}
    \end{subfigure}
    \begin{subfigure}{0.24\linewidth}
        \includegraphics[width=\linewidth]{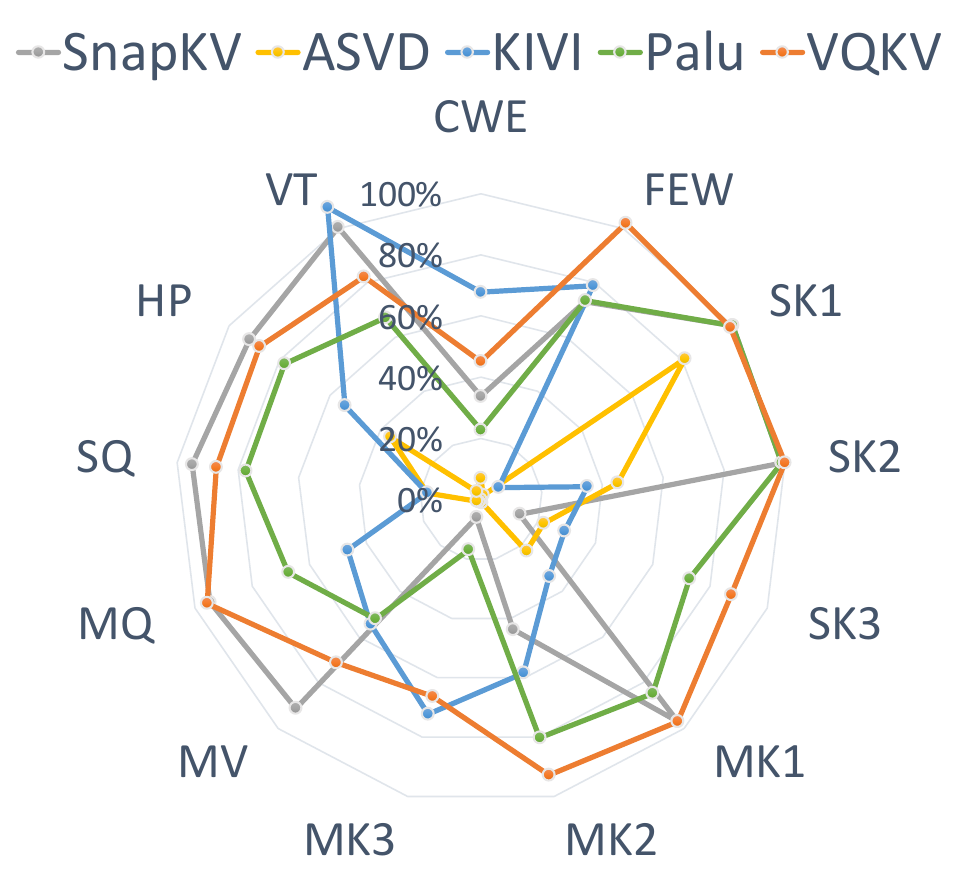}
        \caption{Results on RULER-16K.}
    \end{subfigure}
    \begin{subfigure}{0.24\linewidth}
        \includegraphics[width=\linewidth]{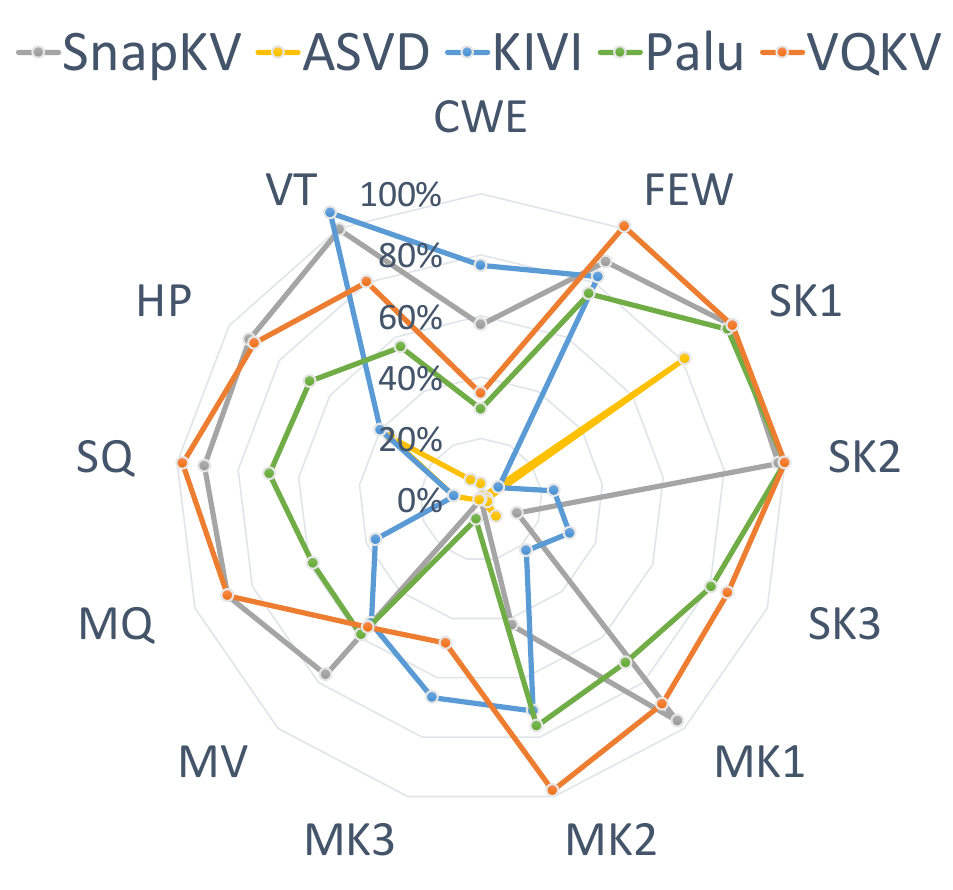}
        \caption{Results on RULER-32K.}
    \end{subfigure}    

    \caption{Detailed results of LLaMA3.2-3B on RULER in different context length. The results of full cache model are taken as the 100\% reference.}
    \label{fig:ruler2}
\end{figure}

\subsection{Long-Context Evaluation}

The long-context evaluation experiments are all conducted on multiple NVIDIA A800 GPUs with FP16 precision. We evaluate our VQKV against other KV cache optimization methods with three long-context benchmarks on OpenCompass \citep{opencompass}: LongBench \citep{longbench}, Needle-In-A-Haystack(NIAH) \citep{niah} and RULER \citep{ruler}. All tasks are set with a truncation context length of 32K. We compare with the ASVD, SnapKV, Palu and KIVI. For fair comparison, we set the ratio parameters to 0.2 in ASVD, keep 2048 recalled middle tokens in SnapKV, retain 70\% of the KV in Palu and apply 4-bit quantization, 75\% compression in KIVI. The results of SnapKV and Palu on NIAH are referred from \citet{fourierkv}.

Across the results on LongBench (Table \ref{tab-mainresult}), NIAH (Figure \ref{fig:niah_llama3.1-8b}, \ref{fig:niah_llama3.2-3b}), and RULER (Figure \ref{fig:ruler1}\ref{fig:ruler2} and Table \ref{tab-mainresult-ruler}), our VQKV consistently achieves the best trade-off between compression and performance. On LongBench, it delivers average scores closest to the uncompressed baseline and surpasses existing baselines under both LLaMA3.1-8B and LLaMA3.2-3B. On NIAH, our approach maintains a perfect 100 score, identical to the full-cache model, while other methods exhibit clear degradation. On RULER, it preserves strong long-context capability, achieving results close to the baseline and substantially outperforming competing methods, even at 32K context length. The overall results collectively demonstrate the robustness and effectiveness of our approach in both general and long-context scenarios.

\begin{figure}
    \centering
    \includegraphics[width=0.7\textwidth]{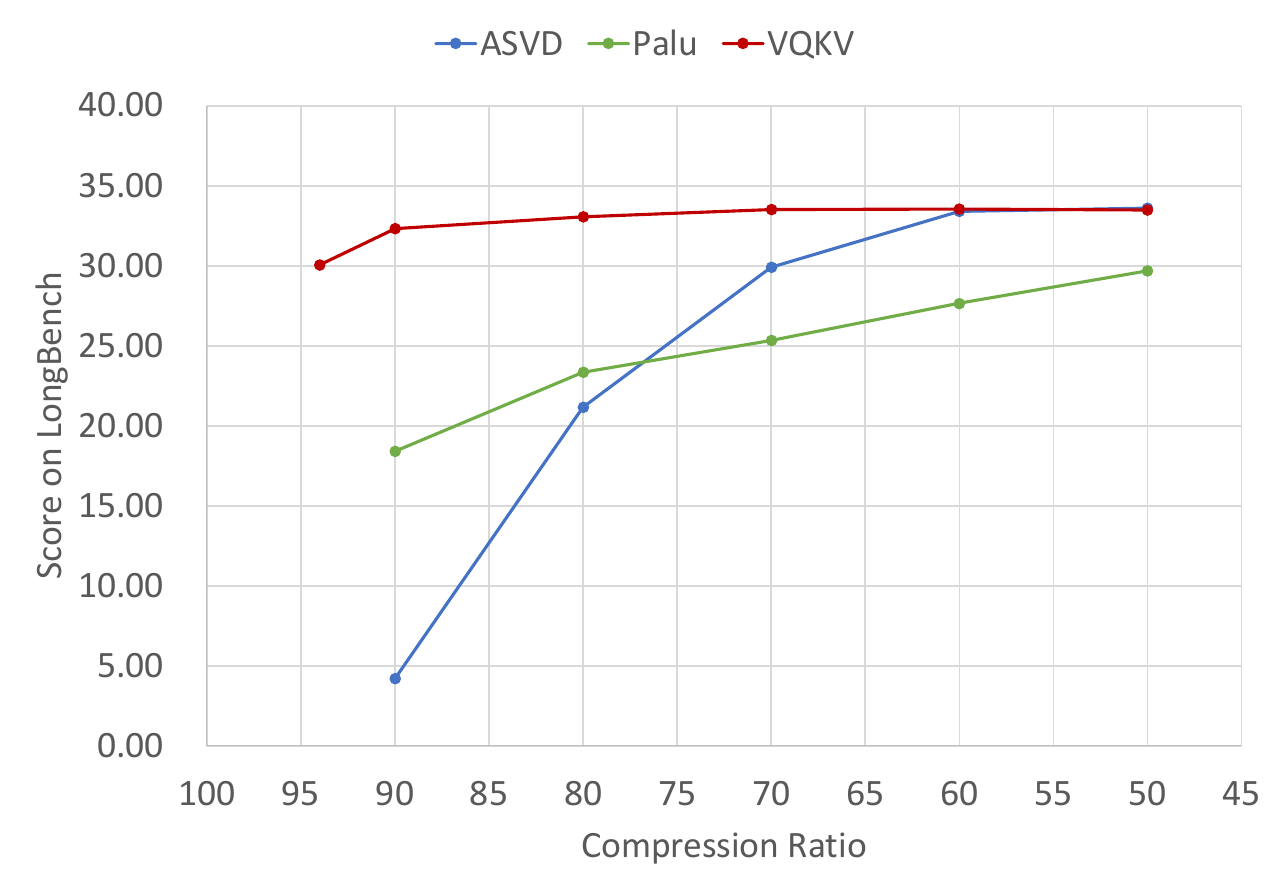}
    \caption{The results of LLaMA3.1-8B on LongBench with different compression ratio.}
    \label{fig:compression-ratio}
\end{figure}

To further demonstrate the effectiveness of VQKV, we compare it with ASVD and Palu under different compression ratios (Table \ref{tab-compressratio}). Figure \ref{fig:compression-ratio} shows that VQKV consistently outperforms these methods across all compression settings and maintains the best performance even at very high compression ratios. These results further highlight the advantages of our method on high compression ratio and high reconstruction fidelity.

\begin{table}[htbp]
    \centering
    \caption{Results of LLaMA3.1-8B and LLaMA3.2-3B on RULER. Our VQKV outperforms other methods on 4K to 32K length.}
    \setlength{\tabcolsep}{4pt}
        \begin{tabular}{@{}l l*{6}{c}@{}}
        \toprule
        & & \textbf{4K} & \textbf{8K} & \textbf{16K} & \textbf{32K} & \textbf{Avg.}\\
        \midrule
        \multirow{5}{*}
        &\textit{LLaMA3.1-8B}  &\gray{94.74} &\gray{92.72} &\gray{93.51} &\gray{90.16} &\gray{92.78} \\
        & + ASVD               & 39.75 & 32.94 & 25.27 & 19.29 & 29.31 \\
        & + SnapKV             & 91.18 & 77.12 & 69.80 & 58.52 & 74.16 \\
        & + Palu               & 74.70 & 66.01 & 59.80 & 52.44 & 63.24 \\
        & + KIVI               & 57.91 & 52.81 & 47.60 & 49.73 & 52.01 \\
        & + \textbf{VQKV(Ours)}& \textbf{92.88} & \textbf{89.61} & \textbf{87.02} & \textbf{79.71} & \textbf{87.31} \\
        \midrule
        \multirow{5}{*}
        &\textit{LLaMA3.2-3B}  &\gray{90.04} &\gray{85.91} &\gray{83.30} &\gray{78.25} &\gray{84.38} \\
        & + ASVD               & 27.15 & 20.83 & 15.95 & 9.51  & 18.36 \\
        & + SnapKV             & 85.33 & 69.57 & 61.73 & 58.74 & 68.84 \\
        & + Palu               & 71.75 & 65.38 & 59.69 & 55.08 & 62.98 \\
        & + KIVI               & 37.92 & 49.37 & 41.41 & 36.28 & 41.25 \\
        & + \textbf{VQKV(Ours)}& \textbf{88.24} & \textbf{78.72} & \textbf{73.58} & \textbf{67.89} & \textbf{77.11} \\
        \bottomrule
    \end{tabular}
    \label{tab-mainresult-ruler}
\end{table}
\begin{table*}[htbp]
    \centering
    \setlength{\tabcolsep}{1.9pt}
    \small
    \caption{Results of LLaMA3.1-8B on LongBench with different compression ratios. The \textbf{Ratio} means discarding ratios. Our VQKV achieves the best performance on comparable compression ratio against other methods.}
    \resizebox{\textwidth}{!}{
    \begin{tabular}{@{}l l*{18}{c}@{}}
        \toprule
          & \multirow{2}{*}{\textbf{Ratio}}
          & \multirow{2}{*}{\textbf{Method}}
          & \multicolumn{3}{c}{\textbf{Single-Doc}}
          & \multicolumn{3}{c}{\textbf{Multi-Doc}}
          & \multicolumn{3}{c}{\textbf{Summary}}
          & \multicolumn{3}{c}{\textbf{Few-shot}}
          & \multicolumn{2}{c}{\textbf{Synthetic}}
          & \multicolumn{2}{c}{\textbf{Code}}
          & \multirow{2}{*}{\textbf{Avg.}} \\
        \cmidrule(lr){4-6} \cmidrule(lr){7-9} \cmidrule(lr){10-12} \cmidrule(lr){13-15} \cmidrule(lr){16-17} \cmidrule(lr){18-19}
        \multicolumn{2}{c}{} &  & \textbf{NQ} & \textbf{Qsp} & \textbf{MF}
        & \textbf{HQ} & \textbf{WQ} & \textbf{Msq}
        & \textbf{GR} & \textbf{QS} & \textbf{MN}
        & \textbf{TR} & \textbf{TQ} & \textbf{SS}
        & \textbf{PC} & \textbf{PR} 
        & \textbf{LCC} & \textbf{Re-P} \\
        \midrule
        \multirow{5}{*}
        &0 & \textit{LLaMA3.1-8B} &\gray{13.0} &\gray{20.4} &\gray{32.1} &\gray{12.0} &\gray{14.0} &\gray{8.7} &\gray{29.7} &\gray{25.2} &\gray{1.0} &\gray{73.5} &\gray{91.0} &\gray{47.2} &\gray{0.8} &\gray{26.8} &\gray{72.0} &\gray{69.2} &\gray{33.5}\\
        \midrule
        & \multirow{3}{*}{50.0\%}
            & ASVD & 11.5 & \textbf{20.3} & 32.0 & 10.6 & 11.8 & 8.5 & 28.4 & 24.3 & 0.1 & 73.5 & 90.7 & 46.0 & \textbf{2.0} & \textbf{40.1} & 70.1 & 67.8 & \textbf{33.6} \\
          & & Palu & \textbf{16.3} & \textbf{20.3} & 31.9 & \textbf{12.1} & 11.8 & 7.7 & 18.1 & 11.6 & \textbf{1.3} & \textbf{76.0} & 90.2 & 35.3 & 1.5 & 8.0 & 67.8 & 65.4 & 29.7 \\
          & & VQKV & 12.6 & 20.1 & \textbf{32.4} & \textbf{12.1} & \textbf{14.1} & \textbf{8.7} & \textbf{29.6} & \textbf{24.9} & 0.9 & 73.5 & \textbf{91.0} & \textbf{47.3} & 0.9 & 26.8 & \textbf{71.7} & \textbf{69.4} & 33.5 \\
        \midrule
        & \multirow{3}{*}{60.0\%}
            & ASVD & 12.7 & 20.0 & \textbf{34.0} & 10.6 & 12.0 & 6.9 & 28.5 & 24.6 & 0.4 & 70.5 & 89.9 & 44.2 & 2.1 & \textbf{40.7} & 69.6 & 67.8 & 33.4 \\
          & & Palu & 10.5 & 18.0 & 25.3 & 11.0 & 12.2 & 7.1 & 17.9 & 8.9 & \textbf{3.8} & \textbf{75.0} & 87.9 & 34.6 & \textbf{2.5} & 5.5 & 62.8 & 59.5 & 27.7 \\
          & & VQKV & \textbf{13.0} & \textbf{20.1} & 32.4 & \textbf{12.2} & \textbf{14.0} & \textbf{8.5} & \textbf{29.6} & \textbf{24.9} & 0.9 & 73.5 & \textbf{91.0} & \textbf{47.3} & 0.8 & 27.3 & \textbf{71.9} & \textbf{69.4} & \textbf{33.6}\\
        \midrule
        & \multirow{3}{*}{70.0\%}
            & ASVD & 5.4 & 19.8 & 20.6 & 9.9 & 11.0 & 7.2 & 23.9 & 23.3 & \textbf{9.1} & 68.0 & 87.9 & 44.2 & 0.8 & 25.4 & 60.3 & 61.7 & 29.9 \\
          & & Palu & 6.2 & 16.9 & 23.6 & 10.8 & 11.9 & 6.1 & 15.9 & 6.5 & 0.8 & \textbf{74.0} & 88.0 & 32.9 & \textbf{1.1} & 6.1 & 50.3 & 54.5 & 25.3 \\
          & & VQKV & \textbf{12.8} & \textbf{20.2} & \textbf{32.5} & \textbf{12.0} & \textbf{13.8} & \textbf{8.6} & \textbf{29.9} & \textbf{24.6} & 0.9 & 73.5 & \textbf{91.0} & \textbf{47.2} & 0.8 & \textbf{27.3} & \textbf{71.9} & \textbf{69.3} & \textbf{33.5} \\
        \midrule
        & \multirow{3}{*}{80.0\%}
            & ASVD & 4.6  & 9.9  & 16.1 & 9.7  & 7.4  & 5.2 & 9.0  & 16.6 & \textbf{12.8} & 60.0 & 78.7 & 33.7 & \textbf{2.8} & 4.9  & 30.4 & 36.9 & 21.2 \\
          & & Palu & 6.4  & 16.6 & 23.1 & 9.7  & 12.3 & 6.6 & 16.5 & 21.5 & 10.7 & 72.5 & 84.6 & 37.1 & 1.3 & 14.3 & 64.7 & 59.1 & 28.6 \\
          & & VQKV & \textbf{14.3} & \textbf{19.8} & \textbf{29.8} & \textbf{11.9} & \textbf{13.8} & \textbf{8.3} & \textbf{26.3} & \textbf{23.9} & 0.9 & \textbf{73.5} & \textbf{90.5} & \textbf{45.7} & 0.8 & \textbf{28.7} & \textbf{71.7} & \textbf{69.5} & \textbf{33.1} \\
        \midrule
        & \multirow{3}{*}{90.0\%}
            & ASVD & 0.4 & 1.7 & 2.0 & 0.7 & 1.0 & 0.6 & 3.4 & 5.0 & \textbf{6.2} & 0.0 & 2.7 & 2.0 & \textbf{2.3} & 2.6 & 19.9 & 16.9 & 4.2 \\
          & & Palu & 3.8 & 13.4 & 19.1 & 8.3 & 8.3 & 4.8 & 11.3 & 4.8 & 4.4 & 64.0 & 67.9 & 10.6 & 2.0 & 4.1 & 31.5 & 36.3 & 18.4 \\
          & & VQKV & \textbf{14.1} & \textbf{18.5} & \textbf{28.9} & \textbf{11.4} & \textbf{14.0} & \textbf{8.0} & \textbf{20.6} & \textbf{23.5} & 0.6 & \textbf{73.0} & \textbf{90.8} & \textbf{45.9} & 0.8 & \textbf{27.7} & \textbf{71.5} & \textbf{68.0} & \textbf{32.3}\\
        \bottomrule
    \end{tabular}
    }
    \label{tab-compressratio}
\end{table*}

\subsection{Memory Efficiency}
\begin{figure}
    \centering
    \includegraphics[width=0.9\textwidth]{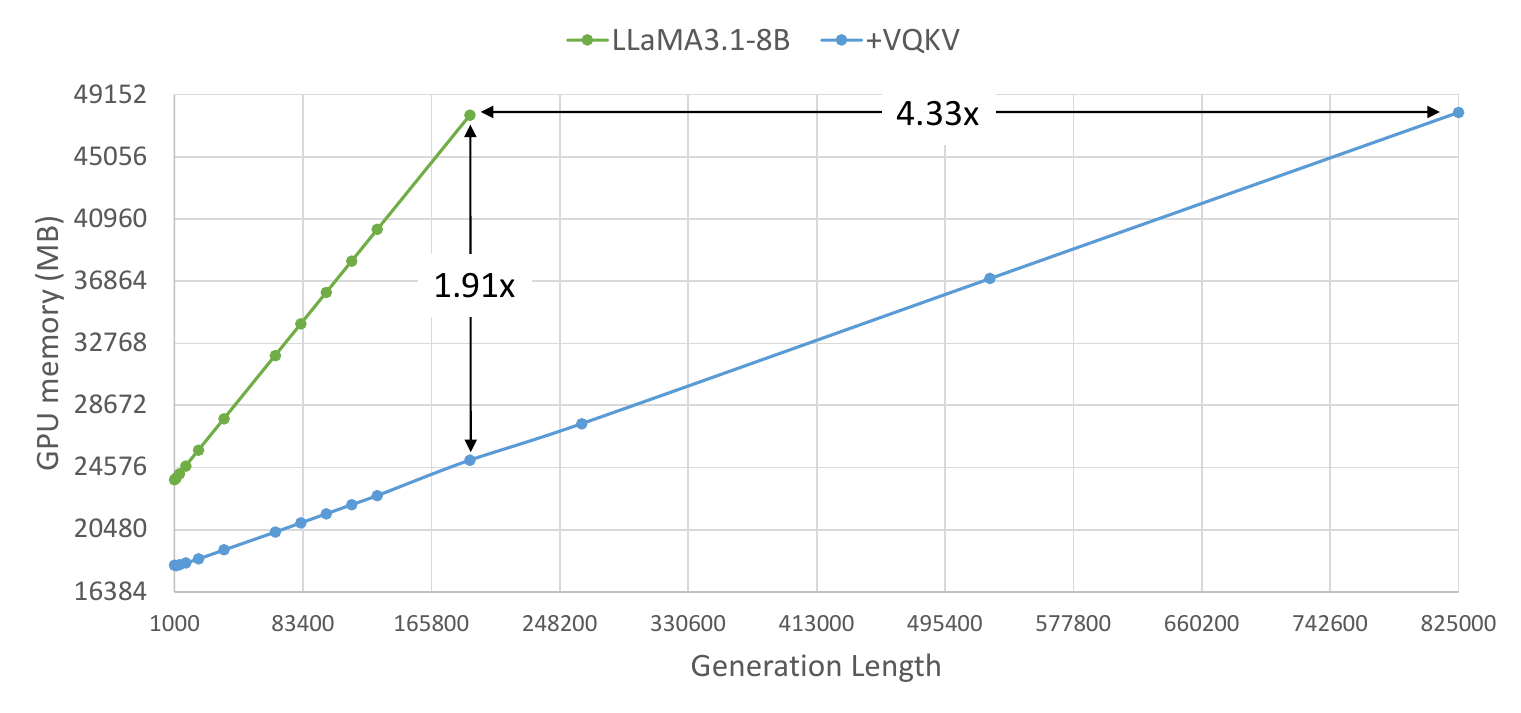}
    \caption{Generation length and corresponding peak memory usage between full-cache LLaMA3.1-8B and our VQKV. The prompt length is 65536 and FlashAttention is used on both.}
    \label{fig-generation}
\end{figure}

To show the memory efficiency of our VQKV, we test the maximum generation length of LLaMA3.1-8B on a single NVIDIA RTX 4090 (48 GB) GPU with FP16 and FlashAttention. As shown in Figure \ref{fig-generation}, our method consistently exhibits substantially lower peak memory usage than the baseline across different generation lengths. When the full-cache baseline model reaches its maximum generation length, VQKV consumes nearly half of the memory required by the baseline. More importantly, our approach dramatically extends the maximum achievable generation length: while the baseline LLaMA3.1-8B model encounters an out-of-memory error at approximately 190k tokens, our method supports sequences exceeding 824k tokens on a single NVIDIA RTX 4090 (48 GB) GPU, representing a more than fourfold increase in usable context length. These results demonstrate that our approach not only preserves strong memory efficiency but also substantially enhances long-sequence generation capability, highlighting its advantages in both compression ratio and memory footprint.

\begin{figure*}
    \centering

    \begin{subfigure}{0.43\linewidth}
        \includegraphics[width=\linewidth]{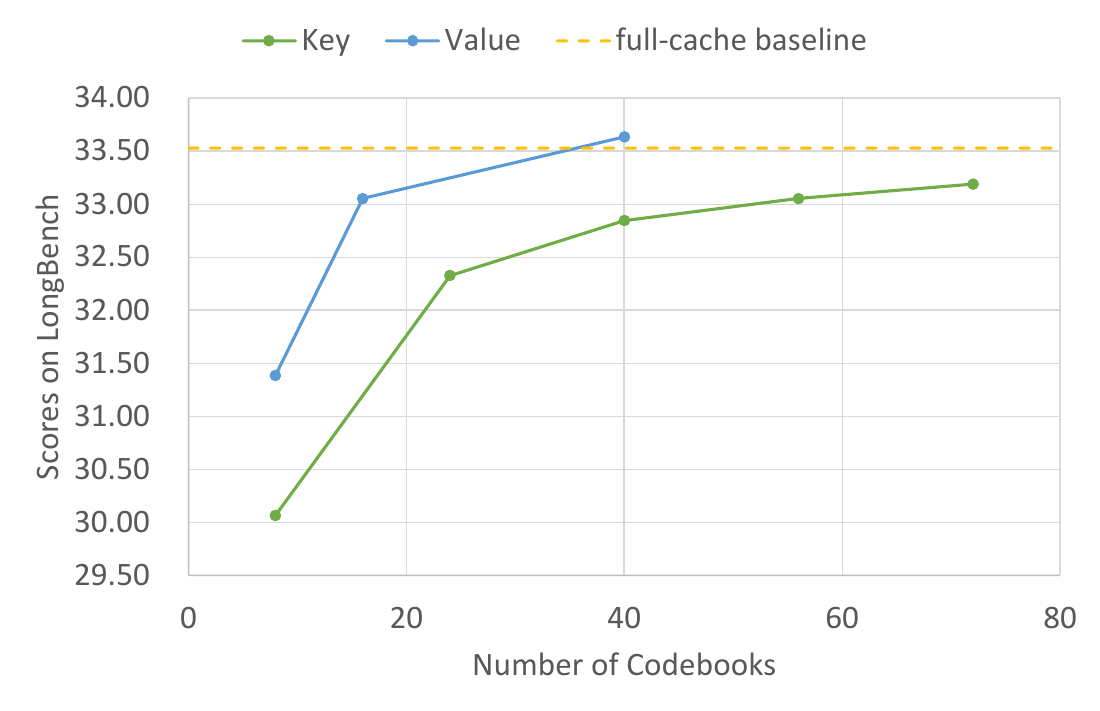}
        \caption{Results of VQKV with different number of codebooks.}
        \label{fig:ablationnumber}
    \end{subfigure}
    \hspace{0.03\linewidth}
    \begin{subfigure}{0.43\linewidth}
        \includegraphics[width=\linewidth]{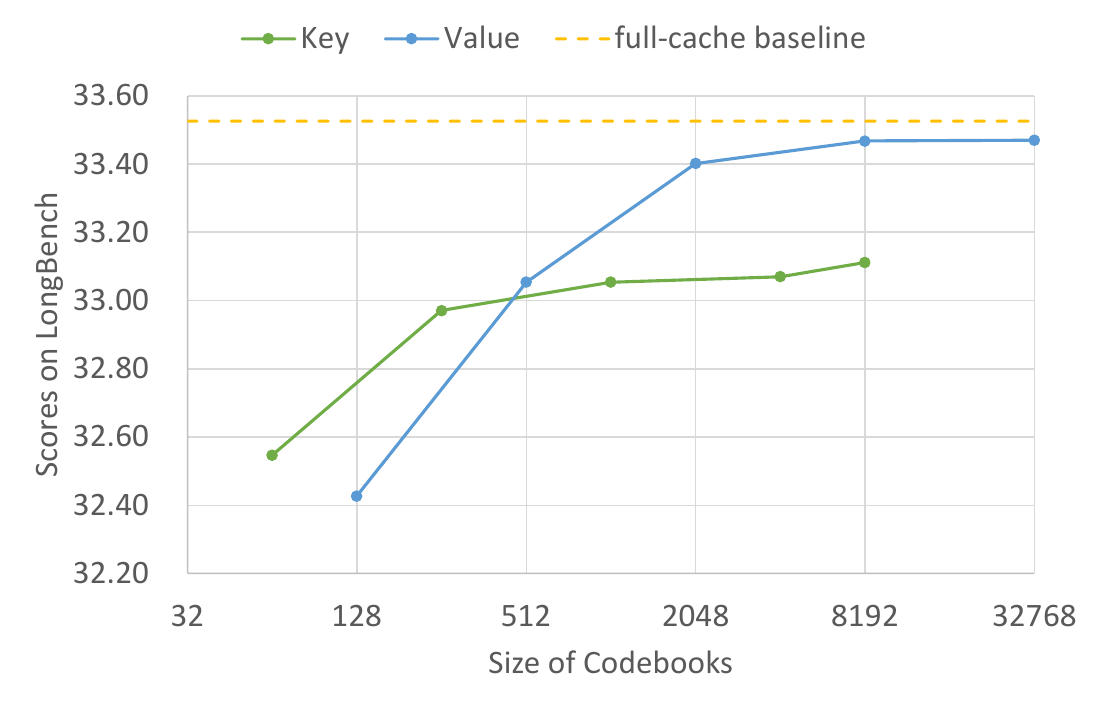}
        \caption{Results of VQKV with different size of codebooks.}
        \label{fig:ablationsize}
    \end{subfigure}
    
    \caption{Results of VQKV on LongBench with different combinations of codebooks on LLaMA3.1-8B.}
    \label{fig:ablation}
\end{figure*}

\subsection{Ablation Study}

We evaluate our VQKV on different combinations of codebook numbers and sizes with LLaMA3.1-8B \citep{llama3.1-8b}.
On Longbench \citep{longbench}, NIAH \citep{niah} and RULER \citep{ruler}, the ablation on codebook numbers show that more and larger codebooks contribute to better performance but lower compression ratio.
The detailed results are available on Appendix \ref{app:detail_result}.

Figure \ref{fig:ablation} analyzes the effects of codebook number and size on LongBench, with separate evaluations for key and value codebooks. 
With a fixed codebook size, increasing the number of codebooks (Figure \ref{fig:ablationnumber}) consistently improves performance, indicating that additional residual stages enhance representation capacity. 
However, the two components exhibit distinct trends: the key codebook performance increases smoothly and gradually saturates, whereas the value codebook performance improves sharply with a small number of codebooks and then quickly plateaus, indicating diminishing returns from adding more quantization stages.

When the number of codebooks is fixed (Figure \ref{fig:ablationsize}), enlarging the codebook size yields performance gains with diminishing returns. The value codebook benefits more at smaller codebook sizes, while the key codebook shows more gradual improvements, and both converge as the codebook size increases, indicating limited benefit from overly large codebooks.

Overall, the value codebook is more sensitive to codebook configurations in both number and size, while the key codebook remains more robust. This observation highlights the importance of fine-grained value representations for long-context modeling and motivates differentiated codebook designs for key and value caches.

\section{Conclusion}
In this paper, we proposed \textbf{VQKV}, a training-free KV cache compression framework that leverages vector quantization to jointly capture correlations within cache vectors. By replacing continuous representations with compact discrete codes, VQKV achieves high compression ratios while preserving fidelity, thus overcoming the limitations of token eviction, feature dimension compression, and scalar quantization. Extensive experiments on LLaMA3.1-8B and LLaMA3.2-3B show that VQKV consistently outperforms existing training-free methods and even surpasses the full-cache baseline in some cases. Moreover, VQKV achieves 4.3× longer generation length on the same memory footprint, demonstrating its effectiveness for memory-efficient long-context inference.

\section{Limitations}
Our method still has room for improvement in decoding efficiency. Although the customized triton kernels have substantially optimized the time efficiency of VQKV, there remains room to further reduce the additional computational overhead introduced by VQ. Such optimizations are not pursued in this paper, and we leave them for future work.


\begin{thebibliography}{39}
\providecommand{\natexlab}[1]{#1}
\providecommand{\url}[1]{\texttt{#1}}
\expandafter\ifx\csname urlstyle\endcsname\relax
  \providecommand{\doi}[1]{doi: #1}\else
  \providecommand{\doi}{doi: \begingroup \urlstyle{rm}\Url}\fi

\bibitem[Baevski et~al.(2020)Baevski, Zhou, Mohamed, and Auli]{wav2vec2}
Alexei Baevski, Yuhao Zhou, Abdelrahman Mohamed, and Michael Auli.
\newblock wav2vec 2.0: A framework for self-supervised learning of speech
  representations.
\newblock \emph{Advances in neural information processing systems},
  33:\penalty0 12449--12460, 2020.

\bibitem[Bai et~al.(2023)Bai, Lv, Zhang, Lyu, Tang, Huang, Du, Liu, Zeng, Hou,
  et~al.]{longbench}
Yushi Bai, Xin Lv, Jiajie Zhang, Hongchang Lyu, Jiankai Tang, Zhidian Huang,
  Zhengxiao Du, Xiao Liu, Aohan Zeng, Lei Hou, et~al.
\newblock Longbench: A bilingual, multitask benchmark for long context
  understanding.
\newblock \emph{arXiv preprint arXiv:2308.14508}, 2023.

\bibitem[Cai et~al.(2024)Cai, Zhang, Gao, Liu, Li, Liu, Lu, Xiong, Dong, Hu,
  et~al.]{pyramidkv}
Zefan Cai, Yichi Zhang, Bofei Gao, Yuliang Liu, Yucheng Li, Tianyu Liu, Keming
  Lu, Wayne Xiong, Yue Dong, Junjie Hu, et~al.
\newblock Pyramidkv: Dynamic kv cache compression based on pyramidal
  information funneling.
\newblock \emph{arXiv preprint arXiv:2406.02069}, 2024.

\bibitem[Chang et~al.(2024)Chang, Lin, Lin, Chen, Hu, Wang, Huang, Ceze,
  Abdelfattah, and Wu]{palu}
Chi-Chih Chang, Wei-Cheng Lin, Chien-Yu Lin, Chong-Yan Chen, Yu-Fang Hu,
  Pei-Shuo Wang, Ning-Chi Huang, Luis Ceze, Mohamed~S Abdelfattah, and
  Kai-Chiang Wu.
\newblock Palu: Compressing kv-cache with low-rank projection.
\newblock \emph{arXiv preprint arXiv:2407.21118}, 2024.

\bibitem[Contributors(2023)]{opencompass}
OpenCompass Contributors.
\newblock Opencompass: A universal evaluation platform for foundation models.
\newblock \url{https://github.com/open-compass/opencompass}, 2023.

\bibitem[Dao(2024)]{flashattention2}
Tri Dao.
\newblock Flash{A}ttention-2: Faster attention with better parallelism and work
  partitioning.
\newblock In \emph{International Conference on Learning Representations
  (ICLR)}, 2024.

\bibitem[Dao et~al.(2022)Dao, Fu, Ermon, Rudra, and R{\'e}]{flashattention}
Tri Dao, Daniel~Y. Fu, Stefano Ermon, Atri Rudra, and Christopher R{\'e}.
\newblock Flash{A}ttention: Fast and memory-efficient exact attention with
  {IO}-awareness.
\newblock In \emph{Advances in Neural Information Processing Systems
  (NeurIPS)}, 2022.

\bibitem[Douze et~al.(2025)Douze, Guzhva, Deng, Johnson, Szilvasy, Mazar{\'e},
  Lomeli, Hosseini, and J{\'e}gou]{faiss}
Matthijs Douze, Alexandr Guzhva, Chengqi Deng, Jeff Johnson, Gergely Szilvasy,
  Pierre-Emmanuel Mazar{\'e}, Maria Lomeli, Lucas Hosseini, and Herv{\'e}
  J{\'e}gou.
\newblock The faiss library.
\newblock \emph{IEEE Transactions on Big Data}, 2025.

\bibitem[Dubey et~al.(2024)Dubey, Jauhri, Pandey, Kadian, Al-Dahle, Letman,
  Mathur, Schelten, Yang, Fan, et~al.]{llama3.1-8b}
Abhimanyu Dubey, Abhinav Jauhri, Abhinav Pandey, Abhishek Kadian, Ahmad
  Al-Dahle, Aiesha Letman, Akhil Mathur, Alan Schelten, Amy Yang, Angela Fan,
  et~al.
\newblock The llama 3 herd of models.
\newblock \emph{arXiv e-prints}, pages arXiv--2407, 2024.

\bibitem[Esser et~al.(2021)Esser, Rombach, and Ommer]{vqgan}
Patrick Esser, Robin Rombach, and Bjorn Ommer.
\newblock Taming transformers for high-resolution image synthesis.
\newblock In \emph{Proceedings of the IEEE/CVF conference on computer vision
  and pattern recognition}, pages 12873--12883, 2021.

\bibitem[Gersho and Gray(2012)]{vq-vanilla}
Allen Gersho and Robert~M Gray.
\newblock \emph{Vector quantization and signal compression}, volume 159.
\newblock Springer Science \& Business Media, 2012.

\bibitem[Gokaslan et~al.(2019)Gokaslan, Cohen, Pavlick, and
  Tellex]{openwebtext}
Aaron Gokaslan, Vanya Cohen, Ellie Pavlick, and Stefanie Tellex.
\newblock Openwebtext corpus.
\newblock \url{http://Skylion007.github.io/OpenWebTextCorpus}, 2019.

\bibitem[He et~al.(2022)He, Chen, Xie, Li, Doll{\'a}r, and Girshick]{mae}
Kaiming He, Xinlei Chen, Saining Xie, Yanghao Li, Piotr Doll{\'a}r, and Ross
  Girshick.
\newblock Masked autoencoders are scalable vision learners.
\newblock In \emph{Proceedings of the IEEE/CVF conference on computer vision
  and pattern recognition}, pages 16000--16009, 2022.

\bibitem[Hooper et~al.(2024)Hooper, Kim, Mohammadzadeh, Mahoney, Shao, Keutzer,
  and Gholami]{KVQuant}
Coleman Hooper, Sehoon Kim, Hiva Mohammadzadeh, Michael~W Mahoney, Yakun~S
  Shao, Kurt Keutzer, and Amir Gholami.
\newblock Kvquant: Towards 10 million context length llm inference with kv
  cache quantization.
\newblock \emph{Advances in Neural Information Processing Systems},
  37:\penalty0 1270--1303, 2024.

\bibitem[Hsieh et~al.(2024)Hsieh, Sun, Kriman, Acharya, Rekesh, Jia, Zhang, and
  Ginsburg]{ruler}
Cheng-Ping Hsieh, Simeng Sun, Samuel Kriman, Shantanu Acharya, Dima Rekesh, Fei
  Jia, Yang Zhang, and Boris Ginsburg.
\newblock Ruler: What's the real context size of your long-context language
  models?
\newblock \emph{arXiv preprint arXiv:2404.06654}, 2024.

\bibitem[Hsu et~al.(2021)Hsu, Bolte, Tsai, Lakhotia, Salakhutdinov, and
  Mohamed]{hubert}
Wei-Ning Hsu, Benjamin Bolte, Yao-Hung~Hubert Tsai, Kushal Lakhotia, Ruslan
  Salakhutdinov, and Abdelrahman Mohamed.
\newblock Hubert: Self-supervised speech representation learning by masked
  prediction of hidden units.
\newblock \emph{IEEE/ACM transactions on audio, speech, and language
  processing}, 29:\penalty0 3451--3460, 2021.

\bibitem[Jegou et~al.(2010)Jegou, Douze, and Schmid]{pqnn}
Herve Jegou, Matthijs Douze, and Cordelia Schmid.
\newblock Product quantization for nearest neighbor search.
\newblock \emph{IEEE transactions on pattern analysis and machine
  intelligence}, 33\penalty0 (1):\penalty0 117--128, 2010.

\bibitem[Johnson et~al.(2019)Johnson, Douze, and J{\'e}gou]{johnson2019billion}
Jeff Johnson, Matthijs Douze, and Herv{\'e} J{\'e}gou.
\newblock Billion-scale similarity search with gpus.
\newblock \emph{IEEE Transactions on Big Data}, 7\penalty0 (3):\penalty0
  535--547, 2019.

\bibitem[Kai et~al.(2025)Kai, Zeng, Wang, Bai, He, Jiang, and
  Lin]{kai2025freqkv}
Jushi Kai, Boyi Zeng, Yixuan Wang, Haoli Bai, Ziwei He, Bo~Jiang, and Zhouhan
  Lin.
\newblock Freqkv: Frequency domain key-value compression for efficient context
  window extension.
\newblock \emph{arXiv preprint arXiv:2505.00570}, 2025.

\bibitem[Li et~al.(2024{\natexlab{a}})Li, Zhang, Liu, and Chen]{niah}
Mo~Li, Songyang Zhang, Yunxin Liu, and Kai Chen.
\newblock Needlebench: Can llms do retrieval and reasoning in 1 million context
  window?, 2024{\natexlab{a}}.
\newblock URL \url{https://arxiv.org/abs/2407.11963}.

\bibitem[Li et~al.(2024{\natexlab{b}})Li, Huang, Yang, Venkitesh, Locatelli,
  Ye, Cai, Lewis, and Chen]{snapkv}
Yuhong Li, Yingbing Huang, Bowen Yang, Bharat Venkitesh, Acyr Locatelli,
  Hanchen Ye, Tianle Cai, Patrick Lewis, and Deming Chen.
\newblock Snapkv: Llm knows what you are looking for before generation.
\newblock \emph{Advances in Neural Information Processing Systems},
  37:\penalty0 22947--22970, 2024{\natexlab{b}}.

\bibitem[Liu et~al.(2024{\natexlab{a}})Liu, Feng, Wang, Wang, Liu, Zhao, Dengr,
  Ruan, Dai, Guo, et~al.]{mla}
Aixin Liu, Bei Feng, Bin Wang, Bingxuan Wang, Bo~Liu, Chenggang Zhao, Chengqi
  Dengr, Chong Ruan, Damai Dai, Daya Guo, et~al.
\newblock Deepseek-v2: A strong, economical, and efficient mixture-of-experts
  language model.
\newblock \emph{arXiv preprint arXiv:2405.04434}, 2024{\natexlab{a}}.

\bibitem[Liu et~al.(2024{\natexlab{b}})Liu, Zhao, Zhuo, Lin, Xin, Li, Qin,
  Qiao, Li, and Gao]{lumina-mgpt}
Dongyang Liu, Shitian Zhao, Le~Zhuo, Weifeng Lin, Yi~Xin, Xinyue Li, Qi~Qin,
  Yu~Qiao, Hongsheng Li, and Peng Gao.
\newblock Lumina-mgpt: Illuminate flexible photorealistic text-to-image
  generation with multimodal generative pretraining.
\newblock \emph{arXiv preprint arXiv:2408.02657}, 2024{\natexlab{b}}.

\bibitem[Liu et~al.(2025{\natexlab{a}})Liu, He, Wang, Li, Song, Liu, Li, Liu,
  Huang, Guo, et~al.]{fourierkv}
Xiaoran Liu, Siyang He, Qiqi Wang, Ruixiao Li, Yuerong Song, Zhigeng Liu,
  Linlin Li, Qun Liu, Zengfeng Huang, Qipeng Guo, et~al.
\newblock Beyond homogeneous attention: Memory-efficient llms via
  fourier-approximated kv cache.
\newblock \emph{arXiv preprint arXiv:2506.11886}, 2025{\natexlab{a}}.

\bibitem[Liu et~al.(2026)Liu, Song, Wang, Ge, Lamb, Guo, Chen, Zhou, and
  Lin]{conceptlm}
Yuliang Liu, Yunchong Song, Yixuan Wang, Kewen Ge, Alex Lamb, Qipeng Guo, Kai
  Chen, Bowen Zhou, and Zhouhan Lin.
\newblock Next concept prediction in discrete latent space leads to stronger
  language models.
\newblock \emph{arXiv preprint arXiv:2602.08984}, 2026.

\bibitem[Liu et~al.(2025{\natexlab{b}})Liu, Luo, Guo, Ni, Zhou, Guan, Guo, Cui,
  Feng, Guo, et~al.]{vq-llm}
Zihan Liu, Xinhao Luo, Junxian Guo, Wentao Ni, Yangjie Zhou, Yue Guan, Cong
  Guo, Weihao Cui, Yu~Feng, Minyi Guo, et~al.
\newblock Vq-llm: High-performance code generation for vector quantization
  augmented llm inference.
\newblock In \emph{2025 IEEE International Symposium on High Performance
  Computer Architecture (HPCA)}, pages 1496--1509. IEEE, 2025{\natexlab{b}}.

\bibitem[Liu et~al.(2024{\natexlab{c}})Liu, Yuan, Jin, Zhong, Xu, Braverman,
  Chen, and Hu]{kivi}
Zirui Liu, Jiayi Yuan, Hongye Jin, Shaochen Zhong, Zhaozhuo Xu, Vladimir
  Braverman, Beidi Chen, and Xia Hu.
\newblock Kivi: A tuning-free asymmetric 2bit quantization for kv cache.
\newblock \emph{arXiv preprint arXiv:2402.02750}, 2024{\natexlab{c}}.

\bibitem[MetaAI(2024)]{llama3.2-3b}
MetaAI.
\newblock Llama 3.2: Revolutionizing edge ai and vision with open, customizable
  models, 2024.

\bibitem[Su et~al.(2024)Su, Ahmed, Lu, Pan, Bo, and Liu]{rope}
Jianlin Su, Murtadha Ahmed, Yu~Lu, Shengfeng Pan, Wen Bo, and Yunfeng Liu.
\newblock Roformer: Enhanced transformer with rotary position embedding.
\newblock \emph{Neurocomputing}, 568:\penalty0 127063, 2024.

\bibitem[Van Den~Oord et~al.(2017)Van Den~Oord, Vinyals, et~al.]{vqvae}
Aaron Van Den~Oord, Oriol Vinyals, et~al.
\newblock Neural discrete representation learning.
\newblock \emph{Advances in neural information processing systems}, 30, 2017.

\bibitem[Wang et~al.(2024)Wang, Li, Ning, Yuan, Yan, Dai, and Wang]{cskv}
Luning Wang, Shiyao Li, Xuefei Ning, Zhihang Yuan, Shengen Yan, Guohao Dai, and
  Yu~Wang.
\newblock Cskv: Training-efficient channel shrinking for kv cache in
  long-context scenarios.
\newblock \emph{arXiv preprint arXiv:2409.10593}, 2024.

\bibitem[Wang et~al.(2025)Wang, Zhu, Liu, Yang, Fang, and He]{vq-vla}
Yating Wang, Haoyi Zhu, Mingyu Liu, Jiange Yang, Hao-Shu Fang, and Tong He.
\newblock Vq-vla: Improving vision-language-action models via scaling
  vector-quantized action tokenizers.
\newblock \emph{arXiv preprint arXiv:2507.01016}, 2025.

\bibitem[Xiao et~al.(2023)Xiao, Tian, Chen, Han, and Lewis]{streamingllm}
Guangxuan Xiao, Yuandong Tian, Beidi Chen, Song Han, and Mike Lewis.
\newblock Efficient streaming language models with attention sinks.
\newblock \emph{arXiv preprint arXiv:2309.17453}, 2023.

\bibitem[Yuan et~al.(2023)Yuan, Shang, Song, Wu, Yan, and Sun]{asvd}
Zhihang Yuan, Yuzhang Shang, Yue Song, Qiang Wu, Yan Yan, and Guangyu Sun.
\newblock Asvd: Activation-aware singular value decomposition for compressing
  large language models.
\newblock \emph{arXiv preprint arXiv:2312.05821}, 2023.

\bibitem[Zeghidour et~al.(2021)Zeghidour, Luebs, Omran, Skoglund, and
  Tagliasacchi]{soundstream}
Neil Zeghidour, Alejandro Luebs, Ahmed Omran, Jan Skoglund, and Marco
  Tagliasacchi.
\newblock Soundstream: An end-to-end neural audio codec.
\newblock \emph{IEEE/ACM Transactions on Audio, Speech, and Language
  Processing}, 30:\penalty0 495--507, 2021.

\bibitem[Zhang et~al.(2023{\natexlab{a}})Zhang, Li, Zhang, Zhan, Wang, Zhou,
  and Qiu]{speechgpt}
Dong Zhang, Shimin Li, Xin Zhang, Jun Zhan, Pengyu Wang, Yaqian Zhou, and
  Xipeng Qiu.
\newblock Speechgpt: Empowering large language models with intrinsic
  cross-modal conversational abilities.
\newblock \emph{arXiv preprint arXiv:2305.11000}, 2023{\natexlab{a}}.

\bibitem[Zhang et~al.(2023{\natexlab{b}})Zhang, Sheng, Zhou, Chen, Zheng, Cai,
  Song, Tian, R{\'e}, Barrett, et~al.]{h2o}
Zhenyu Zhang, Ying Sheng, Tianyi Zhou, Tianlong Chen, Lianmin Zheng, Ruisi Cai,
  Zhao Song, Yuandong Tian, Christopher R{\'e}, Clark Barrett, et~al.
\newblock H2o: Heavy-hitter oracle for efficient generative inference of large
  language models.
\newblock \emph{Advances in Neural Information Processing Systems},
  36:\penalty0 34661--34710, 2023{\natexlab{b}}.

\bibitem[Zheng et~al.(2022)Zheng, Vuong, Cai, and Phung]{movq}
Chuanxia Zheng, Tung-Long Vuong, Jianfei Cai, and Dinh Phung.
\newblock Movq: Modulating quantized vectors for high-fidelity image
  generation.
\newblock \emph{Advances in Neural Information Processing Systems},
  35:\penalty0 23412--23425, 2022.

\bibitem[Zhu et~al.(2024)Zhu, Li, Xin, Xia, and Xu]{simvq}
Yongxin Zhu, Bocheng Li, Yifei Xin, Zhihua Xia, and Linli Xu.
\newblock Addressing representation collapse in vector quantized models with
  one linear layer.
\newblock \emph{arXiv preprint arXiv:2411.02038}, 2024.

\end{thebibliography}
\bibliographystyle{plainnat}

\newpage
\appendix
\section{Detailed Results in Ablation Study}
\label{app:detail_result}
We conduct ablation studies on the number and size of codebooks. On LLaMA3.1-8B \citep{llama3.1-8b}, we evaluate different combinations of codebook numbers and codebook sizes, and test them on LongBench \citep{longbench}, NIAH \citep{niah}, and RULER \citep{ruler}. The detailed experimental results are presented as follows.

\begin{table*}[htbp]
    \centering
    \setlength{\tabcolsep}{2pt}
    \small
    \caption{Detailed results of different combinations of codebooks of VQKV on LLaMA3.1-8B \citep{llama3.1-8b}. \textbf{Ratio} means discarding ratios.}
    \begin{tabular}{cccccccccccc}
        \toprule
        \textbf{$N^k$} & \textbf{$S^k$} & \textbf{$N^v$} & \textbf{$S^v$} & Ratio & \textbf{LongBench} & \textbf{NIAH} & \textbf{RULER} & \textbf{RULER\_4K} & \textbf{RULER\_8K} & \textbf{RULER\_16K} & \textbf{RULER\_32K} \\
        \midrule
        56 & 1024 & 16 & 128 & 83.6\% & 32.43 & 99.86  & 82.70 & 90.58 & 86.28 & 80.58 & 73.35 \\
        \textit{56} & \textit{1024} & \textit{16} & \textit{512} & \textit{82.8\%} & \textit{33.05} & \textit{100.00} & \textit{87.31} & \textit{92.88} & \textit{89.61} & \textit{87.02} & \textit{79.71} \\
        56 & 1024 & 16 & 2048 & 82.0\% & 33.40 & 100.00 & 87.49 & 93.96 & 89.55 & 87.02 & 79.42 \\
        56 & 1024 & 16 & 8192 & 81.3\% & 33.47 & 100.00 & 90.33 & 94.08 & 91.70 & 90.09 & 85.45 \\
        56 & 1024 & 16 & 32768 & 80.5\% & 33.47 & 100.00 & 90.38 & 94.27 & 91.56 & 90.31 & 85.39 \\
        \midrule
        56 & 1024 & 8 & 512 & 84.6\% & 31.39 & 95.94 & 61.16 & 74.43 & 65.60 & 56.85 & 47.78 \\
        \textit{56} & \textit{1024} & \textit{16} & \textit{512} & \textit{82.8\%} & \textit{33.05} & \textit{100.00} & \textit{87.31} & \textit{92.88} & \textit{89.61} & \textit{87.02} & \textit{79.71} \\
        56 & 1024 & 40 & 512 & 77.5\% & 33.64 & 100.00 & 91.36 & 94.45 & 92.03 & 92.01 & 86.96 \\
        \midrule
        56 & 64 & 16 & 512 & 88.3\% & 32.55 & 100.00 & 78.91 & 87.72 & 82.03 & 76.28 & 69.62 \\
        56 & 256 & 16 & 512 & 85.5\% & 32.97 & 100.00 & 85.62 & 92.60 & 88.47 & 84.50 & 76.91 \\
        \textit{56} & \textit{1024} & \textit{16} & \textit{512} & \textit{82.8\%} & \textit{33.05} & \textit{100.00} & \textit{87.31} & \textit{92.88} & \textit{89.61} & \textit{87.02} & \textit{79.71} \\
        56 & 4096 & 16 & 512 & 80.1\% & 33.07 & 100.00 & 87.99 & 93.30 & 89.93 & 87.54 & 81.17 \\
        56 & 8192 & 16 & 512 & 78.7\% & 33.11 & 100.00 & 88.18 & 93.29 & 90.19 & 87.94 & 81.32 \\
        \midrule
        8 & 1024 & 16 & 512 & 94.5\% & 30.07 & 37.51 & 23.71 & 42.94 & 22.56 & 16.61 & 12.71 \\
        24 & 1024 & 16 & 512 & 90.6\% & 32.33 & 99.86 & 73.27 & 84.48 & 76.49 & 71.34 & 60.75 \\
        40 & 1024 & 16 & 512 & 86.7\% & 32.85 & 100.00 & 84.68 & 92.53 & 87.75 & 83.91 & 74.55 \\
        \textit{56} & \textit{1024} & \textit{16} & \textit{512} & \textit{82.8\%} & \textit{33.05} & \textit{100.00} & \textit{87.31} & \textit{92.88} & \textit{89.61} & \textit{87.02} & \textit{79.71} \\
        72 & 1024 & 16 & 512 & 78.9\% & 33.19 & 100.00 & 88.08 & 93.43 & 90.18 & 87.48 & 81.22 \\
        \bottomrule
        
    \end{tabular}
\end{table*}

\end{document}